\documentclass[runningheads]{llncs}
\usepackage[T1]{fontenc}

\usepackage{adjustbox}
\usepackage{amsmath,amssymb,amsfonts}
\usepackage{booktabs}
\usepackage{bm}
\usepackage{graphicx}
\usepackage{hyperref}

\usepackage{subcaption}
\usepackage{textcomp}
\usepackage{xcolor}
\usepackage{xspace}

\DeclareMathAlphabet{\mathsfit}{\encodingdefault}{\sfdefault}{m}{sl}
\SetMathAlphabet{\mathsfit}{bold}{\encodingdefault}{\sfdefault}{bx}{n}
\newcommand{\tens}[1]{\bm{\mathsfit{#1}}}

\newcommand{\ie}{i.\,e.,\xspace}
\newcommand{\eg}{e.\,g.,\xspace}
\newcommand{\iid}{i.\,i.\,d.}

\newcommand{\tXi}{\bm{\Xi}}
\newcommand{\GraphMLP}{Graph-MLP\xspace}
\DeclareMathOperator*{\concat}{\scalebox{1}[1.5]{$\parallel$}}

\definecolor{forestgreen}{rgb}{0, 0.6, 0.32}
\newcommand\good[1]{\textcolor{forestgreen}{#1}}
\newcommand\bad[1]{\textcolor{red}{#1}}
\usepackage{orcidlink}

\begin{document}
\title{The Split Matters: Flat Minima Methods for Improving the Performance of GNNs}
\titlerunning{Flat Minima Methods for GNNs}
\author{Nicolas Lell%
\orcidlink{0000-0002-6079-6480}
\and
Ansgar Scherp
\orcidlink{0000-0002-2653-9245}}
\authorrunning{N. Lell and A. Scherp}
\institute{Universität Ulm, Germany
\email{\{nicolas.lell, ansgar.scherp\}@uni-ulm.de}}
\maketitle              

\begin{abstract}
When training a Neural Network, it is optimized using the available training data with the hope that it generalizes well to new or unseen testing data.
At the same absolute value, a flat minimum in the loss landscape is presumed to generalize better than a sharp minimum.
Methods for determining flat minima have been mostly researched for independent and identically distributed (i.\,i.\,d.) data such as images. 
Graphs are inherently non-i.\,i.\,d. since the vertices are edge-connected. 
We investigate flat minima methods and combinations of those methods for training graph neural networks (GNNs). 
We use GCN and GAT as well as extend Graph-MLP to work with more layers and larger graphs. 
We conduct experiments on small and large citation, co-purchase, and protein datasets with different train-test splits in both the transductive and inductive training procedure.
Results show that flat minima methods can improve the performance of GNN models by over 2 points, if the train-test split is randomized. 
Following Shchur et al., randomized splits are essential for a fair evaluation of GNNs, as other (fixed) splits like ``Planetoid'' are biased. Overall, we provide important insights for improving and fairly evaluating flat minima methods on GNNs.
We recommend practitioners to always use weight averaging techniques, in particular EWA when using early stopping.
While weight averaging techniques are only sometimes the best performing method, they are less sensitive to hyperparameters, need no additional training, and keep the original model unchanged. 
All source code is available under \url{https://github.com/Foisunt/FMMs-in-GNNs}.


\end{abstract}

\section{Introduction}

\begin{figure*}
     \centering
     \begin{subfigure}[b]{0.3\textwidth}
         \centering
         \includegraphics[width=\textwidth]{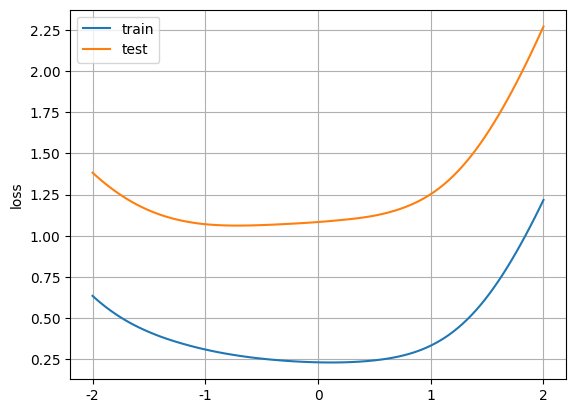}
         \caption{Train and test loss}
         \label{fig:trte}
     \end{subfigure}%
     \begin{subfigure}[b]{0.3\textwidth}
         \centering
         \includegraphics[width=\textwidth]{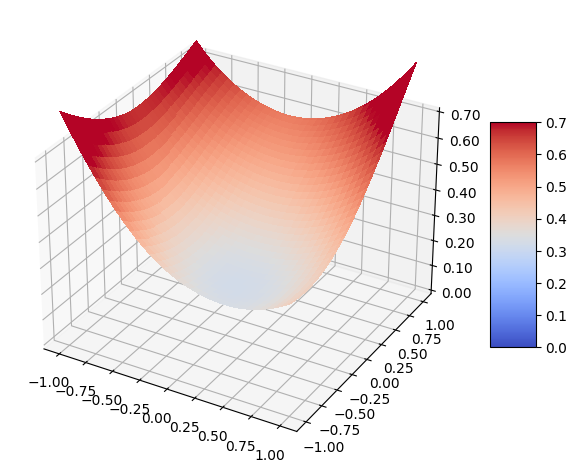}
         \caption{Train loss without a flat minimum method}
         \label{fig:base}
     \end{subfigure}%
     \begin{subfigure}[b]{0.3\textwidth} 
         \centering
         \includegraphics[width=\textwidth]{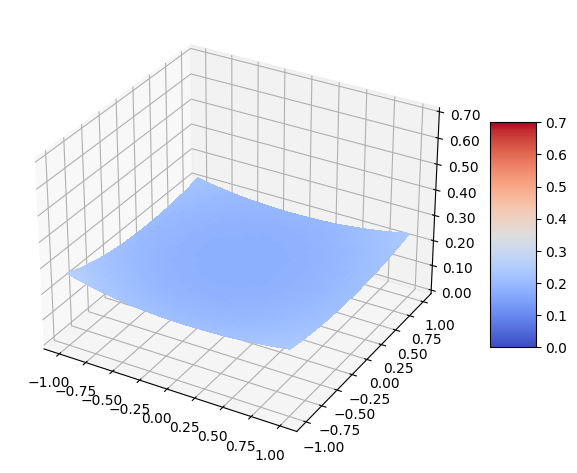}
         \caption{Train loss after training using SAM}
         \label{fig:sam}
     \end{subfigure}
        \caption{Loss of GCN on CiteSeer with the Planetoid split. Plots following \cite{DBLP:conf/nips/Li0TSG18}.}
        \label{fig:three graphs}
\end{figure*}

Flat minima are regions in the weight space of a neural model where the error function remains largely stable.
It is argued that such larger regions of the error function with a constant, low score correspond to less chance of overfitting of the model and thus show higher generalization performance~\cite{HochreiterS94,flatminima}.
We demonstrate this in Figure \ref{fig:trte}, where we plot the training and testing loss of the same model when changing its weights following a random direction.
In that example, the loss landscapes are shifted between train and test data.
Therefore, finding a flat minimum or choosing a central point in a flat region can lead to better generalization compared to a model with the lowest possible loss in a sharper minimum.
Methods for determining flat minima have been researched in the past largely on toy examples and for data that are independent and identically distributed (\iid) such as images, \eg \cite{swa,Foret21sharpness,zhao2022penalizing,GSAM,orvieto2022anticorrelated,SAF}.

Graph neural networks (GNNs) deal with non-i.\,i.\,d. graph data, since vertices are connected via edges.
GNNs are powerful models but are likewise also known to be difficult to train and susceptible to the training procedure~\cite{Shchur18pitfalls}.
Even small changes in the hyperparameters, data split, etc. can lead to unstable training and lack of generalization performance.

We tackle these challenges of GNNs by transferring flat minima methods to graphs.
We consider a wide selection of weight-averaging and sharpness-aware flat minima methods, 
including the well known methods SWA~\cite{swa} and SAM~\cite{Foret21sharpness}, and lesser known or new ones like
Anticorrelated Perturbed Gradient Descent~\cite{orvieto2022anticorrelated}, Penalizing Gradient Norm~\cite{zhao2022penalizing}, and Sharpness Aware Training for Free~\cite{SAF}.
We also apply existing and new combinations such as Penalizing Gradient Norm~\cite{orvieto2022anticorrelated} plus ASAM~\cite{Kwon21asam}.
We evaluate the performance of flat minima methods on different GNN architectures using small and large benchmark datasets.
As GNNs, we use the well known Graph Convolutional Network (GCN)~\cite{Kipf17gcn} and Graph Attention Network (GAT)~\cite{Velickovic18gat} as well as the novel \GraphMLP~\cite{graphMLPwoMP}, which operates without the classical message passing.
Regarding the evaluation procedure, we follow Shchur et al.~\cite{Shchur18pitfalls} who warned that on the common benchmark datasets Cora, CiteSeer, and PubMed the train and test splits heavily impact the models' performance and can lead to an arbitrary reranking of similarly good GNNs.
Thus, in addition to the commonly employed (fixed) ``Planetoid'' split~\cite{planetoid16Yang}, we apply two randomized splits on those datasets.

Our results show that in most cases flat minima methods improve the performance.
But the improvement heavily depends on the used model, dataset, dataset split, and flat minima method.
An illustration of the effect of flat minima methods for GNNs can be seen in Figure~\ref{fig:base} showing the training loss surface of a GNN trained \textit{without} any flat minima method versus Figure~\ref{fig:sam} showing the loss surface of the same model but trained \textit{with} SAM.
To the best of our knowledge, we are the first to systematically transfer and analyze the impact of many flat minima methods to non-\iid\ graph data.
Only Kaddour et al.~\cite{Shchur18pitfalls} applied two flat minima methods SAM and SWA to study images, text, and graphs.
Thus, while their study covers multiple domains, it is limited w.r.t. to the number of minima methods used.
In addition, they only consider fixed train/test splits.
Overall, the contributions of this work are:
\begin{itemize}
    \item 
    We have transferred flat minima methods to operate on non-\iid\ graph data.
    We show that they can improve the performance of GNNs.
    
    \item We perform extensive systematic experiments to measure the influence of flat minima methods depending on the GNN architecture, dataset, and data splits.
    We use both, the transductive as well as inductive training procedure.

   \item We demonstrate that using random splits is essential for fairly evaluating not only GNN models but also the flat minima methods.
   
  \item We combine flat minima methods and show that this improves the performance even further.

\end{itemize}

Below, we review flat minima methods and introduce graph neural networks.
In Section~\ref{sec:methods}, we describe in more detail the flat minima methods used in our experiments.
Section~\ref{sec:expap} introduces the experimental apparatus.
The results are reported in Section~\ref{sec:results} and discussed in Section~\ref{sec:discussion}.

\section{Related Work}
\label{sec:RW}

First, we discuss works in the search of finding flat minima.
Second, we introduce graph neural networks and describe representative models, which we use in our experiments.

\subsection{Searching for Flat Minima}

Hochreiter and Schmidhuber \cite{HochreiterS94,flatminima} were among the first who searched for flat minima in neural networks.
They suggest that finding flatter minima leads to simpler neural networks with better generalization performance.

\paragraph*{SAM-based Approaches}
Foret at al.~\cite{Foret21sharpness} introduced a now popular method that improves generalization through the promotion of flatter minima which they call Sharpness Aware Minimization (SAM).
Their idea is to minimize the loss at the approximated worst (adversarial) point in an explicit region around the model's current parameters and they show that SAM improves the performance and robustness to label noise of Convolutional Neural Networks (CNNs).
SAM showed to also improve the performance of Vision Transformers~\cite{DBLP:conf/iclr/ChenHG22} and Language Models~\cite{Bahri21sharpness}.
Some follow up works used SAM to improve performance on other tasks like model compression~\cite{DBLP:journals/corr/abs-2111-12273,DBLP:journals/corr/abs-2205-12694}.
Other follow up work focused on improving the efficiency of SAM \cite{lookSam}.
For example, Brock et al.~\cite{Brock21high} sampled only a subset of each batch to accelerate the adversarial point calculation.

A different line of follow up work focused on improving the performance of SAM. \label{rw:sharpred}
Kwon et al.~\cite{Kwon21asam} introduced Adaptive SAM (ASAM), which compensates the influence of parameter scaling on the adversarial step.
Kim et al.~\cite{DBLP:conf/icml/Kim0HH22} proposed Fisher SAM which replaces the fixed euclidean balls used by SAM with ellipsoids induced by Fisher information.
Zhao et al.~\cite{zhao2022penalizing} proposed a method which penalizes the gradient norm to find flatter optima and show that SAM is a special case of this method.
Zhuang et al.~\cite{GSAM} proposed Surrogate Gap Guided SAM (GSAM), which in addition to the usual SAM objective, also explicitly minimizes the sharpness.

\paragraph*{Averaging Approaches}
One averaging approach is ensembling~\cite{zhou2012ensemble}, which combines multiple models' outputs to a single, usually more accurate prediction.
For example, Devlin et al.~\cite{bert} showed that an ensemble of BERT large models gain roughly $1\%$ F1 score on SQuAD 1.1 compared to a single model.
\cite{snapshot} proposed a method called Snapshot Ensemble, which averages a single model's predictions at different points during training.

There are also averaging approaches other than ensembling.
Izmailov et al.~\cite{swa} proposed the now well known method Stochastic Weight Averaging (SWA) which averages a single model's weights at different points during training.
There are some follow up works on SWA, for example using SWA in low precision training can close the performance difference, even when using only 8 bits for each parameter and gradient\cite{DBLP:conf/icml/YangZKBWS19}.
Recently, Wortsman et al.~\cite{soup} showed that most of the good models obtained during hyperparameter tuning lie in the same flat region and that averaging those models' weights leads to better performance compared to simply using the best found model.
Further extension and uses of SWA are described by \cite{DBLP:conf/iclr/GuptaSD20,swaRe}.

\paragraph*{Other Approaches}
Perturbed Gradient Descent (PGD) is a version of gradient descent where noise is injected in every epoch. 
This helped to escape from local minima~\cite{pgd_localminima} and saddle points~\cite{pgd_saddle}.
Orvieto et al.~\cite{orvieto2022anticorrelated} proposed a modification of PGD, which they call Anticorrelated PGD (Anti-PGD).
The idea of Anti-PGD is to inject noise in the current epoch, depending on the noise injected in the previous epoch.
They prove for some special problems that this leads the optimizer to the widest optimum and show that it increases performance on benchmark datasets.
Damian et al.~\cite{DBLP:conf/nips/DamianML21} showed that adding noise to the labels when using Stochastic Gradient Descent (SGD) leads to flatter optima and better generalization.

Du et al.~\cite{SAF} proposed a method which they coin Sharpness-Aware Training for Free (SAF).
They consider SAM's adversarial point approximation as too costly, and instead rely on a trajectory loss to reduce sharpness.

\subsection{Graph Neural Networks}

Graph Neural Networks (GNNs) are neural networks that are designed to work with graph data.
That means in addition to the vertex features, a GNN also uses the adjacency information which connects different data points.
In the following, the adjacency matrix is denoted with $\bm{A}$, the normalized adjacency matrix with $\bm{\hat{A}} = \bm{D}^{-1/2} \bm{A} \bm{D}^{-1/2}$ with $\bm{D}_{ii}=\sum_j \bm{A}_{ij}$ and layer $l$'s output with $\bm{H}^{(l)}$.
Many GNNs follow the message passing architecture~\cite{Hamilton2020graph,DBLP:journals/aiopen/ZhouCHZYLWLS20}, where the current vertex aggregates the features of neighboring vertices to update its own feature vector.
Different aggregation and update methods then lead to different GNNs.
A well known example is the Graph Convolution Network (GCN)~\cite{Kipf17gcn}, where the implementations are inspired by CNNs.
In each layer, every vertex combines its neighbors' features to calculate its output as follows: 
$\bm{H}^{(l+1)} = \sigma(\bm{\hat{A}}\bm{H}^{(l)}\bm{W}^{(l)})$.

Another well known GNN is the Graph Attention Network (GAT)~\cite{Velickovic18gat}, which uses attention to weigh each neighboring vertex's importance for the current vertex.
The attention weights are calculated by $a_{ij} = \mathrm{softmax}(\mathrm{FF}(\bm{W} {\bm{h}}_i,\bm{W} {\bm{h}}_j))$, where $\mathrm{FF}$ is a one layer feed forward network.
Those are then used to calculate vertex $i$'s output with:
\begin{equation}
{\bm{h}}^{(l+1)}_i = \concat^K_{k=1}\sigma \left( \sum_{j\in N_i}a^k_{ij}\bm{W}^k{\bm{h}}_j \right)    
\end{equation}
where $\concat$ is the concatenation, $K$ the number of attention heads, and $N_i$ the $1$-hop neighborhood of vertex $i$ including itself.

These GNNs usually use the whole graph in a single batch, \ie require to load the full graph at once.
This makes it difficult to apply GCN and GAT to very large graphs.
There are different methods to scale GNNs that sample subgraphs and train on those instead of the full graph \cite{ChenZS18,graphSage,graphsaint-iclr20}.
Wu et al.~\cite{Wu19sgcn} propose Simplified GCN, which uses only a single message passing layer with the adjacency matrix to some power instead of multiple iterations with the normal adjacency matrix.

A common issue with the GNNs mentioned so far is over-smoothing~\cite{Hamilton2020graph}, which means that after multiple message passing steps, all vertex representations tend to be very similar.
This can either be avoided by restricting the GNNs to usually only one to three layers or adding residual or skip connections to the model.
The Jumping Knowledge model~\cite{jkn} uses skip connections from every layer to the last layer.
Chen et al.~\cite{gcnII} introduced GCNII which utilizes skip connections from the input layer to every hidden layer.
Both ideas make it possible to gain performance by increasing the model depth up to $64$ layers.

\GraphMLP \cite{graphMLPwoMP} is a GNN approach that is not based on the message-passing architecture.
Rather, \GraphMLP employs a standard Multi Layer Perceptron (MLP) on the vertex features and uses a contrastive loss function on the $r$-th power normalized adjacency matrix $\bm{\hat{A}}^r$.
The neighbor contrastive (NC) loss for vertex $i$ is calculated as
\begin{equation}
l_i = -\mathrm{log}\frac{\sum_{j\ne i} \bm{\hat{A}}^r_{ij} \mathrm{exp}(\mathrm{cos}({\bm{z}}_i, {\bm{z}}_j)\tau)}{\sum_{k\ne i} \mathrm{exp}(\mathrm{cos}({\bm{z}}_i, {\bm{z}}_k)\tau)}    
\end{equation}
where ${\bm{z}}_i$ is the embedding/intermediate layer output of vertex $i$ and $\tau$ is a temperature parameter.
Other than GCN and GAT that are full batch by default, \GraphMLP randomly samples a batch from the input graph each epoch.

\section{Flat Minima Methods}
\label{sec:methods}

Here we give a brief introduction for a high level understanding as well as the modified parameter update rules for the flat minima methods used in our work.
For details, we refer to the primary literature.
In the following, we denote the learning rate with $\eta$ and a model's weights as $\tens{W}$, \eg for a $l$-layer GAT $\tens{W} = [\bm{W}^{(1)}, \mathrm{FF}^{(1)}, ..., \bm{W}^{(l)}, \mathrm{FF}^{(l)}]$.
We begin with SAM and works extending SAM, followed by weight-averaging methods.
Finally, we discuss SAF and Anti-PGD.

SAM~\cite{Foret21sharpness} searches for a model that has a region with low loss around it, instead of finding the model with lowest loss.
SAM minimizes the loss of the approximately worst point $\tens{W}_{adv}$ in the region of size $\rho$ around the model.
The adversarial point is approximated via $\tens{W}_{adv} = \tens{W}_n + \rho (\nabla L(\tens{W}_n))/(||\nabla L(\tens{W}_n)||_2)$ and is used for the model's training by $\tens{W}_{n+1} = \tens{W}_n - \eta \nabla L(\tens{W}_{adv})$.

ASAM~\cite{Kwon21asam} considers that a model's parameters can be scaled without changing the loss.
By incorporating the weights' norms into the parameter update, the performance of SAM can be increased.
Formally, ASAM changes SAM's calculation to 
$\tens{W}_{adv} = \tens{W}_n + \rho (T^2_{\tens{W}} \nabla L(\tens{W}_n))/(||T_{\tens{W}} \nabla L(\tens{W}_n)||_2)$ with $T_{\tens{W}}$ being a normalization operator for the weights.

PGN: The gradient's norm directly corresponds to the sharpness of the model's current weights.
By penalizing the gradient norm (PGN) during training, the models tend to reach flatter optima~\cite{zhao2022penalizing}.
PGN generalizes SAM with the update rule $\tens{W}_{n+1} = \tens{W}_n - \eta ((1-\alpha)\nabla L(\tens{W})+\alpha \nabla L(\tens{W}_{adv}))$, where $\alpha$ is a new balancing parameter.
We also experiment with a combination of PGN with ASAM, where we use ASAM to calculate $\tens{W}_{adv}$, which we call \textbf{PGNA}.

GSAM: SAM only optimizes the worst point in a region around it.
But it might be better to explicitly minimize the sharpness of said region as well.
GSAM~\cite{GSAM} does this by adding a sharpness term to the loss while ensuring that the gradient of the sharpness term does not increase SAM's original loss via an orthogonal projection.
This results in $\tens{W}_{n+1} = \tens{W}_n - \eta (\nabla L(\tens{W}_{adv}) - \alpha \nabla L(\tens{W})_{\perp})$, where $\alpha$ is a balancing parameter.
We also use a variant called \textbf{GASAM}, which uses ASAM to calculate $\tens{W}_{adv}$.

SWA~\cite{swa} is based on the observation that models trained using SGD with cyclic or high constant learning rates tend to traverse flat regions of the loss.
As the loss landscapes are slightly shifted between training, validation, and test data, which can be seen in Figure \ref{fig:trte}, the center point of the training loss basin should generalize best.
To exploit this assumption, SWA calculates an average model $\tens{W}_\text{swa}$ by proportionally adding the current weights every $k$-th epoch by $\tens{W}_\text{swa} = (\tens{W}_\text{swa} \cdot n_\text{models} + \tens{W}_\text{current})/(n_\text{models}+1)$, with $n_\text{models}$ being the number of models averaged. 
As we use early stopping following Shchur et al.~\cite{Shchur18pitfalls}, we do not know in advance for how many epochs each model trains.
Thus, different to the original SWA which used predefined compute budgets, we start averaging at epoch $begin$ and stop averaging $end$ epochs after early stopping triggered.

EWA: 
Pre-experiments showed that the number of epochs a model trains heavily depend on the GNN architecture, dataset, split, and smoothing method used.
In our case, it ranges from $5$ epochs (GCN on CiteSeer with the 622 split) to about $2000$ epochs (\GraphMLP on arXiv).
This makes it hard to choose the $begin$ parameter as one ideally only wants to average models that are already close to the optimum.
Therefore, we also experiment with exponential weight averaging (EWA), \ie $\tens{W}_\text{ewa} = \alpha \tens{W}_\text{ewa} + (1-\alpha) \cdot \tens{W}_\text{current}$.
With the introduction of the new hyperparameter $\alpha$, we expect that EWA works well independent of the number of training epochs.

Anti-PGD: Noise can be injected into gradient descent to improve the training through faster escape from saddle points or local minima.
When the loss is in a valley, anti-correlated noise additionally moves the model to a wider section of the valley \cite{orvieto2022anticorrelated}.
The model's weights are updated by $\tens{W}_{n+1} = \tens{W}_n-\eta\nabla L(\tens{W}_{n+1})+(\tXi_{n+1}-\tXi_n)$, where $\tXi_i$ is a random tensor with variance $\sigma^2$.
After training the model for some epochs with noise injection, the noise injection is stopped for the remaining training to improve convergence of the model.

SAF: Since SAM computes two gradients, it uses about twice the time per weight update compared to standard SGD.
As mentioned in Section~\ref{rw:sharpred}, there are some methods to reduce the impact of computing the additional gradient, but SAF~\cite{SAF} removes the second gradient calculation all together.
Instead it approximates the sharpness by the change of the model output over the epochs.
Specifically, a new trajectory loss $L^{tra} = \lambda/|B| \cdot \sum_{i\in B} \mathrm{KL}({\bm{y}}^{(e-E)}_i/\tau, {\bm{y}}^{(e)}_i/\tau)$ is added to the normal loss.
In that case $\mathrm{KL()}$ is the Kullback–Leibler divergence, $B$ a batch, ${\bm{y}}^{(e)}$ is the model's output at the current epoch $e$, ${\bm{y}}^{(e-E)}$ is the model's output $E$ epochs ago, $\tau$ is a temperature, and $\lambda$ the loss weight.

\section{Experimental Apparatus}
\label{sec:expap}
In this section, we present our experimental apparatus, \ie the used datasets, models, procedure, and measures.

\subsection{Datasets}
\label{sec:ds}

We use different benchmark datasets to evaluate the flat minima methods.
Table~\ref{tab:datasets} reports statistics of the datasets.
Cora~\cite{CoraCiteseer}, CiteSeer~\cite{CoraCiteseer}, PubMed~\cite{pubmed}, and OGB arXiv~\cite{ogb} are citation graphs. 
Amazon Computers and Amazon Photo~\cite{Shchur18pitfalls} are co-purchase graphs.
For these datasets the task is single label vertex classification.
Protein Protein Interaction (PPI)~\cite{PPI} is a collection of 24 protein graphs with 20 of those used for training and 2 each for validation and testing.
The task for PPI is multi label vertex classification.
There are $121$ different labels with each vertex having between $0$ and $101$ labels, an average of $36.9\pm 22.2$ labels.

The ``Planetoid'' (in tables ``plan'') train-test split~\cite{planetoid16Yang} is often used for Cora, CiteSeer, and PubMed.
It is a fixed split with $20$ vertices per class for training, $500$ vertices for validation, and $1000$ for testing.
Shchur et al.~\cite{Shchur18pitfalls} showed that changing the train-test split can arbitrarily rerank GNN methods of similar performance.
Thus, we also use two other kinds of randomly generated splits that we also use for the Computers and Photo datasets.
The random Planetoid ``ra-pl'' split follows \cite{Shchur18pitfalls} with $20$ vertices per class for training, $30$ per class for validation, and all other vertices for testing.
The ``622'' split, for example used in \cite{gcnII}, consists of $60\%$ of the vertices for training, $20\%$ for validation, and $20\%$ for testing.

For OGB arXiv, we use the default training (paper before 2018), validation (paper from 2018), and test split (paper after 2018).
We add reverse and self edges, making the graph essentially undirected, which is needed for good performance.
This increases the number of edges from $1\,166\,243$ to $2\,484\,941$.
Note that reverse edges are already included by default in the other benchmark datasets.
Table~\ref{tab:ds_splits} summarizes all used splits.

OGB arXiv is used in the transductive setting, \ie all vertex features and edges are available during training.
PPI is used in the inductive setting, \ie no validation and test vertices and edges are used during training.
For the other dataset we use both settings.
However, we do not use the ra-pl split in the inductive setting.
The reason is that the induced subgraph over the 20 vertices drawn per class in the ra-pl split typically results in no connected vertices.
Thus, there are no edges in the subgraph for training, which renders this split ineffective.

\begin{table}[!ht]
    \centering
    \caption{Datasets used. $C$ is the number of classes and $F$ is the feature size.
    As PPI contains multiple graphs, the sum of vertices and edges is shown here. $\dagger$Number after adding self and reverse edges; number before is $1\,166\,243$.}
    \label{tab:datasets}
\small
    \begin{tabular}{l|r|r|r|r}\toprule
         Dataset  & $C$ & $F$ & $|V|$ & $|E|$ \\ \midrule
         Cora     &  $7$ & $1\,433$ &   $2\,708$ &    $10\,556$\\
         CiteSeer &  $6$ & $3\,703$ &   $3\,327$ &     $9\,104$\\
         PubMed   &  $3$ &   $500$ &  $19\,717$ &    $88\,648$\\\midrule
         Computers &  $10$ &   $767$ &  $13\,752$ &    $491\,722$\\
         Photo    &  $8$ &   $745$ &  $7\,650$ &    $238\,162$\\\midrule
         arXiv    & $40$ &   $128$ & $169\,343$ & $\dagger2\,484\,941$\\ 
         PPI      & $121$ &   $50$ & $56\,944$ & $1\,587\,264$\\ \bottomrule
    \end{tabular}
\end{table}

\begin{table}[!ht]
    \centering
    \caption{Dataset splits. Besides the fixed ``Planetoid'' split, we use: ``ra-pl'' denotes random splits as used by \cite{Shchur18pitfalls} and ``622'' denotes random $60\%$ train, $20\%$ validation, and $20\%$ test split.}
    \label{tab:ds_splits}
\small    
    \begin{tabular}{l|l|r|r|r}\toprule
          & Split type & Train  & Val     & Test  \\ \midrule
         Cora    & plan    &  $140$ &   $500$&  $1\,000$\\
                 &ra-pl&  $140$ &   $210$&  $2\,358$\\
                 & 622   & $1\,621$&   $542$&   $545$\\
         CiteSeer& plan    &  $120$ &   $500$&  $1\,000$\\
                 &ra-pl&  $120$ &   $180$&  $3\,027$\\
                 & 622   & $1\,993$&   $666$&   $668$\\
         PubMed  & plan    &   $60$ &   $500$& $1\,000$\\
                 &ra-pl&   $60$ &    $90$&$19\,567$\\
                 & 622   &$11\,829$& $3\,944$& $3\,944$\\ \midrule
        Computers& ra-pl & $200$ & $300$ & $13\,252$\\
                 & 622 & $8\,246$ & $2\,750$ & $2\,756$ \\
        Photo   & ra-pl & $160$ & $240$ & $7\,250$\\
                 & 622 & $4\,586$ & $1\,530$ & $1\,534$\\ \midrule
         arXiv &default&$90\,941$&$29\,799$&$48\,603$\\ 
       PPI       &default& $44\,906$ & $6\,514$ & $5\,524$\\ \bottomrule
    \end{tabular}
\end{table}

\subsection{Procedure}
\label{sec:proc}

We precompute the random splits of our datasets (ra-pl, 622) such that they are consistent between models and methods.
PPI is used inductively, \ie only training vertices and edges connecting those are used for training. 
arXiv is used transductively, \ie labeled training vertices are available together with the other vertices but without labels.
We use both setups for the other datasets.
The actual experimental procedure is then executed in two steps.
First, we optimize the GNN models (GCN, GAT, \GraphMLP) in a traditional way without any flat minima methods as described in Section~\ref{sec:hyper}.
Second, using the hyperparameters fixed in the first step, we we add the flat minima methods and only optimize their respective hyperparameters(again described in Section~\ref{sec:hyper}).
For both hyperparameter searches, only the training and validation sets are used.
Subsequently, we evaluate the models.
Additionally, we combine promising flat minima methods (without further hyperparameter tuning) on a subset of the datasets.
We run multiple repeats with different seeds for each of our experiments.
For the smaller datasets, we use 100 repeats, and 10 repeats for the arXiv and PPI datasets.
We report the mean performance of the GNN models averaged over those repeats.
For the flat minima methods, we report the difference to the respective GNN model.
This allows for fast visual assessment of the results.
In addition, we the report the standard deviation over the different runs for all models.

\subsection{Hyperparameters}
\label{sec:hyper}
We tune the GNN hyperparameters, fix them, and then tune the flat minima methods' hyperparameters.
We optimized all hyperparameters individually per setting (in-, transductive), dataset, and split.
In summary, we use early stopping with a patience of 100 epochs, two to three layer models with a hidden size smaller or equal to 256 for the small datasets and slightly modify \GraphMLP.
For PPI and arXiv we use deeper (up to ten layers) and wider models that also use residual connections.
For the flat minima methods we tune each methods' hyperparameters while keeping the base ones from fixed.
For PGN and GSAM we reuse $\rho$ we found for SAM and ASAM.
For details and all final values see \ref{appx:hyper}.

\subsection{Metrics}
For the multi-label PPI dataset, we report weighted Macro-F1 scores.
The F1 score is calculated per class and averaged with weights based on the support of each class.
For all other, single-label datasets we report accuracy.

\section{Results}
\label{sec:results}

The results of the transductive experiments are shown in Table~\ref{tab:trans} with standard deviations shown in Table~\ref{tab:transSD}. 
Regarding the base models, we see that on Cora GAT performs best.
On CiteSeer, PubMed, and Photo, \GraphMLP beats the message passing methods by $1$ to $3$ points.
On Computers \GraphMLP is the best model when using the 622 split but the worst model when using the ra-pl split.
On arXiv GCN performs best with a $0.3$ point lead over GAT.
Regarding the different splits, we can see that compared to the Planetoid split the performance is lower on the ra-pl and higher on the 622 split.
Regarding the flat minima methods, we observe that there is no method that always works best.
The largest improvement is over $2$ points on the CiteSeer ra-pl split with GAT+EWA.
On arXiv all non-weight averaging methods improve the performance of GCN, but only SAF for GAT.
For \GraphMLP on arXiv, EWA improves the performance by $0.82$ points.
There are also some bad combinations.
For example, ASAM reduces the performance of GAT in most cases.

\setlength{\tabcolsep}{1.4pt}
\begin{table*}[!ht]
    \centering
    
    \caption{Transductive mean accuracy per split and dataset. Note: The minima method PGNA is a combination of PGN+ASAM. GASAM combines GSAM and ASAM.
    For the SD over the 100 and 10 runs, we refer to Table \ref{tab:transSD}.}
    \label{tab:trans}
    \scriptsize
    \resizebox{0.99\textwidth}{!}{

    \begin{tabular}{l|r r r | r r r | r r r | r r | r r | r}\toprule
    Dataset   & \multicolumn{3}{c|}{Cora} & \multicolumn{3}{c|}{CiteSeer} & \multicolumn{3}{c|}{PubMed} & \multicolumn{2}{c|}{Computer} & \multicolumn{2}{c|}{Photo} & arXiv \\
    Split     & plan & ra-pl & 622 & plan & ra-pl & 622 & plan & ra-pl & 622 & ra-pl & 622 & ra-pl & 622 & - \\ \midrule
   GCN & $82.02$ & $79.82$ & $88.44$ & $71.39$ & $67.41$ & $76.81$ & $79.34$ & $77.27$ & $89.46$ & $82.78$ & $91.88$ & $90.89$ & $94.55$ & $72.95$\\   \hline
 +SAM & \good{$+0.21$} & \good{$+0.53$} & \good{$+0.05$} & \good{$\mathbf{+1.34}$} & \good{$+1.41$} & \good{$+0.02$} & \bad{$-0.27$} & \good{$+0.24$} & \bad{$-0.13$} & \bad{$-0.07$} & \good{$+0.17$} & \good{$+0.40$} & \bad{$-0.02$} & \good{$+0.10$}\\
 +ASAM & \good{$+0.28$} & \good{$+0.53$} & \good{$+0.02$} & \good{$+0.93$} & \good{$+1.47$} & \bad{$-0.18$} & \good{$+0.18$} & \good{$+0.41$} & \bad{$-0.28$} & \bad{$-0.20$} & \good{$+0.11$} & \good{$+0.30$} & \bad{$-0.03$} & \good{$+0.01$}\\
 +PGN & \good{$+0.04$} & \good{$+0.40$} & \bad{$-0.01$} & \good{$+1.08$} & \good{$+1.91$} & \bad{$-0.01$} & \bad{$-0.06$} & \bad{$-0.09$} & \bad{$-0.07$} & \good{$+0.08$} & \good{$+0.09$} & \good{$+0.37$} & \good{$+0.00$} & \good{$+0.07$}\\
 +PGNA & \good{$\mathbf{+0.35}$} & \good{$+0.70$} & \bad{$-0.01$} & \good{$+0.93$} & \good{$+1.78$} & \bad{$-0.12$} & \good{$\mathbf{+0.22}$} & \good{$+0.33$} & \bad{$-0.02$} & \good{$+0.11$} & \good{$+0.06$} & \good{$+0.28$} & \good{$+0.02$} & \good{$+0.01$}\\
 +GSAM & \good{$+0.15$} & \good{$+0.50$} & \bad{$-0.01$} & \good{$+1.25$} & \good{$+1.55$} & \bad{$-0.01$} & \bad{$-0.19$} & \good{$+0.14$} & \bad{$-0.09$} & \bad{$-0.05$} & \good{$+0.16$} & \good{$+0.38$} & \bad{$-0.00$} & \good{$+0.09$} \\
 +GASAM & \good{$\mathbf{+0.35}$} & \good{$\mathbf{+0.84}$} & \good{$+0.02$} & \good{$+1.16$} & \good{$+1.58$} & \bad{$-0.12$} & \good{$+0.13$} & \good{$\mathbf{+0.43}$} & \bad{$-0.01$} & \bad{$-0.35$} & \good{$+0.10$} & \good{$+0.38$} & \bad{$-0.02$} & \good{$+0.05$}\\  \hline
 +SWA & \bad{$-0.11$} & \good{$+0.39$} & \good{$\mathbf{+0.12}$} & \good{$+0.52$} & \good{$+1.59$} & \good{$+0.03$} & \bad{$-0.60$} & \bad{$-0.48$} & \bad{$-0.09$} & \bad{$-0.59$} & \good{$+0.23$} & \good{$+0.21$} & \good{$\mathbf{+0.09}$} & \bad{$-0.61$} \\
 +EWA & \bad{$-0.09$} & \bad{$-0.01$} & \good{$+0.04$} & \good{$+0.25$} & \good{$\mathbf{+2.09}$} & \good{$\mathbf{+0.26}$} & \good{$+0.04$} & \bad{$-0.09$} & \good{$\mathbf{+0.21}$} & \bad{$-0.36$} & \good{$\mathbf{+0.33}$} & \good{$+0.02$} & \good{$+0.06$} & \bad{$-0.21$} \\ \hline
 +Anti-PGD & \bad{$-0.02$} & \good{$+0.51$} & \good{$+0.05$} & \bad{$-0.04$} & \good{$+1.11$} & \bad{$-0.08$} & \bad{$-0.01$} & \bad{$-0.08$} & \good{$+0.19$} & \good{$+0.17$} & \good{$+0.09$} & \good{$+0.05$} & \bad{$-0.00$} & \good{$\mathbf{+0.13}$} \\
 +SAF & \good{$+0.26$} & \good{$+0.57$} & \bad{$-0.00$} & \good{$+0.47$} & \good{$+0.13$} & \good{$+0.03$} & \good{$+0.01$} & \bad{$-0.00$} & \good{$+0.02$} & \good{$\mathbf{+1.13}$} & \good{$+0.19$} & \good{$\mathbf{+0.59}$} & \bad{$-0.02$} & \good{$+0.11$}\\

 \hline 
\midrule

  GAT & $82.94$ & $80.73$ & $88.42$ & $71.39$ & $69.96$ & $76.55$ & $79.09$ & $77.22$ & $88.59$ & $83.02$ & $92.17$ & $90.56$ & $94.72$ & $72.65$\\
  \hline
 +SAM & \bad{$-0.28$} & \bad{$-0.29$} & \good{$+0.19$} & \good{$+0.07$} & \bad{$-0.06$} & \good{$+0.10$} & \good{$+0.30$} & \good{$+0.09$} & \bad{$-0.11$} & \bad{$-0.28$} & \good{$\mathbf{+0.29}$} & \bad{$-0.15$} & \good{$+0.01$} & \bad{$-0.07$} \\
 +ASAM & \bad{$-0.74$} & \bad{$-0.14$} & \good{$+0.06$} & \bad{$-0.27$} & \bad{$-0.61$} & \bad{$-0.19$} & \bad{$-0.24$} & \bad{$-0.35$} & \bad{$-0.16$} &  \bad{$-0.42$} & \good{$+0.22$} & \bad{$-0.16$} & \good{$+0.08$} & \bad{$-0.03$} \\
 +PGN & \good{$\mathbf{+0.32}$} & \bad{$-0.00$} & \good{$\mathbf{+0.24}$} & \good{$\mathbf{+0.64}$} & \good{$+0.23$} & \good{$+0.12$} & \good{$+0.50$} & \good{$+0.38$} & \good{$\mathbf{+0.09}$} & \good{$\mathbf{+0.27}$} & \good{$+0.23$} & \good{$+0.08$} & \good{$\mathbf{+0.17}$} & \bad{$-0.03$} \\
 +PGNA & \good{$+0.30$} & \bad{$-0.01$} & \good{$+0.20$} & \bad{$-0.22$} & \good{$+0.20$} & \good{$+0.03$} & \good{$+0.29$} & \good{$+0.30$} & \bad{$-0.09$} & \bad{$-0.21$} & \good{$+0.21$} & \good{$+0.06$} & \good{$+0.10$} & \bad{$-0.07$} \\
 +GSAM & \good{$+0.07$} & \bad{$-0.29$} & \good{$+0.17$} & \good{$+0.33$} & \bad{$-0.01$} & \good{$+0.14$} & \good{$\mathbf{+0.88}$} & \good{$+0.09$} & \good{$+0.06$} & \bad{$-0.21$} & \good{$+0.20$} & \bad{$-0.09$} & \good{$+0.11$} &  \bad{$-0.03$} \\
 +GASAM & \bad{$-0.14$} & \good{$\mathbf{+0.02}$} & \bad{$-0.00$} & \bad{$-0.33$} & \bad{$-0.25$} & \good{$+0.05$} & \good{$+0.51$} & \good{$\mathbf{+0.46}$} & \good{$\mathbf{+0.09}$} & \bad{$-0.40$} & \good{$+0.22$} & \bad{$-0.14$} & \good{$+0.10$} &  \bad{$-0.04$} \\
 \hline
 +SWA & \bad{$-0.87$} & \bad{$-0.28$} & \good{$\mathbf{+0.24}$} & \bad{$-0.76$} & \good{$+0.55$} & \good{$+0.19$} & \bad{$-0.87$} & \bad{$-1.25$} & \bad{$-0.09$} & \bad{$-0.77$} & \good{$+0.14$} & \good{$\mathbf{+0.14}$} & \good{$+0.07$} & \bad{$-35.14$} \\
 +EWA & \bad{$-0.26$} & \bad{$-0.06$} & \bad{$-0.05$} & \bad{$-0.31$} & \good{$+0.16$} & \good{$\mathbf{+0.24}$} & \bad{$-0.26$} & \bad{$-0.28$} & \good{$+0.04$} & \bad{$-0.41$} & \good{$+0.23$} & \good{$+\mathbf{0.14}$} & \good{$+0.07$} & \bad{$-41.32$} \\
  \hline
 +Anti-PGD & \good{$+0.01$} & \bad{$-0.06$} & \good{$+0.08$} & \good{$+0.02$} & \good{$\mathbf{+0.59}$} & \bad{$-0.04$} & \good{$+0.15$} & \good{$+0.11$} & \good{$+0.06$} & \bad{$-0.13$} & \good{$+0.03$} & \good{$+0.09$} & \good{$+0.02$} & \bad{$-0.02$} \\
 +SAF & \bad{$-0.13$} & \bad{$-0.08$} & \bad{$-0.06$} & \bad{$-0.05$} & \bad{$-0.01$} & \bad{$-0.09$} & \good{$+0.01$} & \bad{$-0.11$} & \bad{$-0.05$} & \good{$+0.00$} & \good{$+0.05$} & \good{$+0.10$} & \bad{$-0.01$} & \good{$\mathbf{+0.12}$} \\
  \hline
 
\midrule
   \GraphMLP & $80.58$ & $78.76$ & $88.08$ & $74.53$ & $71.36$ & $77.69$ & $82.16$ & $78.19$ & $90.31$ & $81.59$ & $92.25$ & $91.30$ & $95.94$ & $67.79$ \\
  \hline
 +SAM & \good{$\mathbf{+0.63}$} & \good{$+0.45$} & \good{$+0.22$} & \bad{$-0.27$} & \good{$+0.26$} & \good{$+0.06$} & \good{$+0.35$} & \good{$+0.61$} & \bad{$-0.01$} & \bad{$-0.14$} & \good{$+0.04$} & \good{$+0.45$} & \good{$+0.01$} & \good{$+0.77$} \\
 +ASAM & \good{$+0.19$} & \bad{$-0.07$} & \good{$+0.16$} & \bad{$-0.58$} & \good{$+0.14$} & \good{$+0.05$} & \good{$\mathbf{+0.44}$} & \good{$+0.54$} & \good{$+0.04$} & \bad{$-0.18$} & \good{$+0.02$} & \good{$+0.39$} & \good{$+0.07$} & \good{$+0.62$} \\
 +PGN & \good{$+0.40$} & \good{$+0.45$} & \good{$+0.13$} & \good{$\mathbf{+0.20}$} & \good{$+0.17$} & \good{$+0.08$} & \good{$+0.05$} & \bad{$-0.13$} & \good{$+0.05$} & \good{$+0.23$} & \good{$+0.01$} &  \good{$+0.49$} & \good{$+0.06$} &  \good{$+0.75$} \\
 +PGNA & \good{$+0.27$} & \good{$+0.13$} & \good{$+0.11$} & \bad{$-0.15$} & \good{$+0.09$} & \bad{$-0.03$} & \good{$+0.19$} & \good{$+0.15$} & \good{$+0.09$} & \good{$+0.10$} & \good{$+0.05$} & \good{$+0.47$} & \good{$+0.04$} & \good{$+0.79$} \\
 +GSAM & \good{$+0.52$} & \good{$+0.49$} & \good{$\mathbf{+0.27}$} & \bad{$-0.11$} & \good{$+0.20$} & \bad{$-0.06$} & \good{$+0.36$} & \good{$\mathbf{+0.75}$} & \bad{$-0.01$} & \bad{$-0.25$} & \good{$+0.00$} & \good{$+0.32$} & \good{$+0.02$} & \good{$+0.77$} \\
 +GASAM & \good{$+0.20$} & \good{$+0.07$} & \good{$+0.16$} & \bad{$-0.39$} & \good{$+0.29$} & \bad{$-0.11$} & \good{$+0.28$} & \good{$+0.38$} & \good{$+0.04$} & \bad{$-0.19$} & \good{$+0.02$} & \good{$+0.25$} & \good{$\mathbf{+0.08}$} & \good{$+0.68$} \\ \hline
 +SWA & \good{$+0.34$} & \bad{$-0.01$} & \good{$\mathbf{+0.27}$} & \good{$+0.06$} & \good{$+0.51$} & \good{$\mathbf{+0.30}$} & \bad{$-0.45$} & \good{$+0.12$} & \bad{$-0.33$} & \good{$\mathbf{+0.32}$} & \good{$+0.02$} & \good{$\mathbf{+0.67}$} & \good{$+0.06$} & \bad{$-3.44$} \\
 +EWA & \bad{$-0.05$} & \good{$+0.02$} & \good{$+0.01$} & \bad{$-0.01$} & \good{$+0.06$} & \good{$+0.03$} & \bad{$-0.16$} & \good{$+0.07$} & \good{$+0.14$} & \good{$+0.07$} & \good{$+0.10$} & \good{$+0.02$} & \bad{$-0.01$} & \good{$\mathbf{+0.82}$} \\  \hline
 +Anti-PGD & \good{$+0.17$} & \bad{$-0.07$} & \good{$+0.10$} & \good{$+0.02$} & \good{$+0.03$} & \good{$+0.08$} & \bad{$-0.20$} & \bad{$-0.32$} & \good{$+0.14$} & \good{$+0.17$} & \good{$+0.07$} & \good{$+0.22$} & \good{$+0.02$} & \good{$+0.22$} \\
 +SAF & \bad{$-0.06$} & \good{$\mathbf{+1.35}$} & \good{$+0.07$} & \bad{$-0.02$} & \good{$\mathbf{+1.01}$} & \good{$+0.08$} & \good{$+0.02$} & \bad{$-0.11$} & \good{$\mathbf{+0.33}$} & \good{$+0.23$} & \good{$\mathbf{+0.13}$} & \good{$+0.08$} & \good{$+0.07$} & \bad{$-0.06$} \\

    \bottomrule
    \end{tabular}}
\end{table*}

\begin{table*}[!ht]
    \centering
        \caption{Inductive mean accuracy per split and dataset. For PPI we report weighted Macro-F1 scores. Note: The minima method PGNA is a combination of PGN+ASAM. GASAM combines GSAM and ASAM. For the SD over the 100 and 10 runs, we refer to Table \ref{tab:indSD}.}
    \label{tab:ind}
   \footnotesize
       \resizebox{0.8\textwidth}{!}{

    \begin{tabular}{l|r r | r r | r r | r | r | r }\toprule
    Dataset       & \multicolumn{2}{c|}{Cora} & \multicolumn{2}{c|}{CiteSeer} & \multicolumn{2}{c|}{PubMed} & Comp.. & Photo & PPI\\
    Split         & plan & 622 & plan & 622 & plan & 622 & 622 & 622 & -\\ \midrule

  GCN & $81.08$ & $88.02$ & $71.56$ & $76.80$ & $78.63$ & $89.81$ & $91.63$ & $94.49$ & $99.34$\\
 + SAM   & \bad{$-0.39$} & \good{$+0.36$} & \good{$+0.82$} & \good{$+0.33$} & \good{$+0.47$} & \bad{$-0.22$} & \good{$+0.17$} & \good{$+0.03$} & \bad{$-0.01$}\\
 + ASAM  & \bad{$-0.31$} & \good{$+0.37$} & \good{$+0.33$} & \good{$+0.13$} & \good{$+0.45$} & \bad{$-0.44$} & \good{$+0.15$} & \good{$+0.02$} & \good{$+0.00$}\\
 + PGN   & \bad{$-0.04$} & \good{$+0.21$} & \good{$+0.77$} & \good{$+0.19$} & \good{$+0.47$} & \bad{$-0.15$} & \good{$+0.02$} & \good{$+0.06$} & \bad{$-0.01$}\\
 + PGNA  & \good{$+0.02$} & \good{$+0.41$} & \good{$+0.37$} & \good{$+0.26$} & \good{$\mathbf{+0.52}$} & \bad{$-0.06$} & \good{$+0.02$} & \good{$+0.03$} & \bad{$-0.01$}\\
 + GSAM  & \good{$+0.24$} & \good{$+0.21$} & \good{$\mathbf{+1.06}$} & \good{$+0.41$} & \good{$\mathbf{+0.52}$} & \bad{$-0.23$} & \good{$+0.11$} & \good{$+0.03$} & \bad{$-0.00$}\\
 + GASAM & \good{$\mathbf{+0.37}$} & \good{$+0.48$} & \good{$+0.50$} & \good{$+0.29$} & \good{$+0.47$} & \bad{$-0.15$} & \good{$+0.11$} & \good{$+0.01$} & \bad{$-0.00$}\\ \hline
 + SWA   & \bad{$-0.10$} & \good{$\mathbf{+0.50}$} & \bad{$-0.34$} & \good{$+0.34$} & \bad{$-1.06$} & \bad{$-0.47$} & \good{$+0.11$} & \good{$+0.06$} & \bad{$-0.20$}\\
 + EWA   & \bad{$-0.05$} & \good{$+0.38$} & \good{$+0.55$} & \good{$+0.36$} & \bad{$-0.48$} & \good{$\mathbf{+0.17}$} & \good{$\mathbf{+0.37}$} & \good{$+0.04$} & \good{$+0.01$}\\ \hline
 + Anti-PGD   & \bad{$-0.01$} & \good{$+0.37$} & \good{$+0.44$} & \good{$\mathbf{+0.45}$} & \bad{$-0.09$} & \good{$+0.10$} & \good{$+0.05$} & \good{$\mathbf{+0.09}$} & \bad{$-0.01$} \\
 + SAF   & \bad{$-0.01$} & \good{$+0.07$} & \bad{$-0.13$} & \bad{$-0.05$} & \bad{$-0.07$} & \bad{$-0.01$} & \good{$+0.07$} & \bad{$-0.04$} & \good{$\mathbf{+0.04}$}\\

\hline
\midrule

  GAT & $82.20$ & $87.92$ & $72.10$ & $76.69$ & $78.30$ & $88.55$ & $91.66$ & $94.45$ & $99.29$\\
 + SAM & \bad{$-0.16$} & \good{$+0.21$} & \bad{$-0.33$} & \bad{$-0.18$} & \good{$+0.31$} & \good{$+0.06$} & \good{$+0.38$} & \good{$+0.00$} & \good{$+0.04$}\\
 + ASAM & \bad{$-0.45$} & \good{$+0.14$} & \bad{$-0.36$} & \bad{$-0.06$} & \bad{$-0.03$} & \bad{$-0.16$} & \good{$+0.30$} & \good{$+0.16$} & \good{$+0.04$}\\
 + PGN & \good{$+0.23$} & \good{$+0.17$} & \good{$+0.03$} & \good{$+0.04$} & \bad{$-0.10$} & \good{$+0.00$} & \good{$+0.31$} & \good{$+0.15$} & \good{$\mathbf{+0.05}$}\\
 + PGNA & \bad{$-0.24$} & \good{$+0.22$} & \good{$\mathbf{+0.18}$} & \good{$+0.01$} & \bad{$-0.26$} & \bad{$-0.03$} & \good{$+0.30$} & \good{$\mathbf{+0.19}$} & \good{$\mathbf{+0.05}$}\\
 + GSAM & \bad{$-0.15$} & \good{$+0.21$} & \bad{$-0.33$} & \bad{$-0.24$} & \good{$\mathbf{+0.38}$} & \good{$+0.11$} & \good{$+0.37$} & \good{$+0.11$} & \good{$+0.04$}\\
 + GASAM & \bad{$-0.30$} & \good{$+0.18$} & \bad{$-0.36$} & \good{$+0.00$} & \bad{$-0.02$} & \bad{$-0.03$} & \good{$+0.33$} & \good{$+0.15$} & \good{$\mathbf{+0.05}$}\\ \hline
 + SWA & \bad{$-1.33$} & \good{$\mathbf{+0.45}$} & \good{$+0.04$} & \good{$\mathbf{+0.29}$} & \bad{$-0.69$} & \good{$+0.00$} & \good{$+0.31$} & \good{$+0.13$} & \bad{$-0.49$}\\
 + EWA & \bad{$-0.04$} & \good{$+0.20$} & \good{$+0.03$} & \good{$\mathbf{+0.29}$} & \bad{$-0.11$} & \good{$\mathbf{+0.19}$} & \good{$\mathbf{+0.40}$} & \good{$+0.08$} & \good{$+0.03$}\\ \hline
 + Anti-PGD & \good{$\mathbf{+0.32}$} & \bad{$-0.01$} & \good{$+0.14$} & \good{$+0.00$} & \bad{$-0.16$} & \good{$+0.04$} & \good{$+0.08$} & \good{$+0.04$} & \good{$+0.01$}\\
 + SAF & \good{$+0.16$} & \good{$+0.08$} & \good{$+0.00$} & \good{$+0.01$} & \good{$+0.03$} & \good{$+0.05$} & \good{$+0.18$} & \good{$+0.05$} & \bad{$-0.15$} \\

\hline
\midrule

  \GraphMLP & $68.72$ & $77.47$ & $69.37$ & $74.13$ & $81.20$ & $89.51$ & $87.39$ & $92.87$ & $54.63$\\ \hline
 + SAM & \good{$+2.25$} & \good{$\mathbf{+0.45}$} & \bad{$-0.23$} & \good{$+0.18$} & \bad{$-0.12$} & \good{$+0.13$} & \good{$\mathbf{+0.58}$} & \good{$\mathbf{+0.51}$} & \bad{$-3.35$}\\
 + ASAM & \good{$+1.06$} & \bad{$-0.27$} & \bad{$-0.36$} & \bad{$-0.13$} & \bad{$-0.06$} & \bad{$-0.12$} & \bad{$-0.01$} & \good{$+0.15$} & \bad{$-1.18$}\\
 + PGN & \good{$+2.83$} & \good{$+0.27$} & \bad{$-0.07$} & \good{$+0.38$} & \good{$+0.05$} & \good{$\mathbf{+0.35}$} & \good{$+0.42$} & \good{$+0.37$} & \good{$+1.81$}\\
 + PGNA & \good{$+1.83$} & \bad{$-0.19$} & \bad{$-0.07$} & \bad{$-0.04$} & \bad{$-0.01$} & \good{$+0.09$} & \good{$+0.09$} & \good{$+0.15$} & \good{$+1.26$}\\
 + GSAM & \good{$+2.00$} & \good{$+0.36$} & \bad{$-0.12$} & \good{$+0.26$} & \bad{$-0.08$} & \good{$+0.31$} & \good{$+0.55$} & \good{$+0.45$} & \bad{$-3.20$}\\
 + GASAM & \good{$+2.49$} & \good{$+0.43$} & \bad{$-0.23$} & \good{$+0.45$} & \good{$+0.06$} & \bad{$-0.03$} & \bad{$-0.01$} & \good{$+0.22$} & \bad{$-0.58$}\\ \hline
 + SWA & \bad{$-0.91$} & \good{$+0.32$} & \bad{$-0.10$} & \good{$+0.18$} & \bad{$-0.65$} & \bad{$-0.29$} & \good{$+0.25$} & \good{$+0.40$} & \bad{$-2.23$}\\
 + EWA & \bad{$-0.94$} & \good{$+0.00$} & \bad{$-0.01$} & \good{$+0.03$} & \bad{$-0.06$} & \good{$+0.16$} & \good{$+0.03$} & \good{$+0.18$} & \bad{$-0.94$}\\ \hline
 + Anti-PGD & \good{$\mathbf{+2.86}$} & \bad{$-0.01$} & \bad{$-0.09$} & \good{$+0.02$} & \good{$+0.03$} & \good{$+0.31$} & \good{$+0.09$} & \good{$+0.02$} & \good{$+1.15$}\\
 + SAF & \good{$+1.75$} & \good{$+0.41$} & \good{$\mathbf{+1.08}$} & \good{$\mathbf{+0.51}$} & \good{$\mathbf{+0.19}$} & \good{$+0.14$} & \good{$+0.06$} & \good{$+0.00$} & \good{$\mathbf{+2.62}$}\\

   \bottomrule
    \end{tabular}}
\end{table*}

Table~\ref{tab:ind} shows the results for the inductive experiments with standard deviations shown in Table~\ref{tab:indSD}.
As explained in Section \ref{sec:ds}, the ra-pl split was not used here.
The performance of most models lies within $1$ point of the performance in the transductive setting.
\GraphMLP drops over $10$ points on Cora and around $4$ points on CiteSeer, Computers, and Photo, but is still the best model for PubMed.
On PPI, GCN is the best model while \GraphMLP has a $45$ point drop in F1 score.
Regarding the flat minima methods, the overall picture is the same.
In many cases the best performing method from the transductive setting still is one of the best ones in the inductive setting.
For example, for GCN on Cora, PubMed, and Computers, the same flat minima method works best in both settings.
In other cases it changes, for example for \GraphMLP on CiteSeer with the Planetoid split now only SAF increases the performance, while in the transductive setting PGN workes best and SAF reduces the performance.
On Computers and Photo nearly all flat minima methods improve the performance.
On PPI SAF works for GCN and all SAM variants improve GAT.
For \GraphMLP on PPI, the effects of the minima methods are mixed. 
For example, SAM reduces the performance by $3.35$ points, PGN increases it by $1.81$, and SAF brings a large improvement with $2.62$ points.

\section{Discussion}
\label{sec:discussion}

\subsection{Key Insights}
Regarding the flat minima methods, we can see that there is no method that always works best.
However, for each combination of a base GNN model and dataset, there is at least one flat minima method that improves the performance.
But in many cases some flat minima methods also reduce the performance.
For GCN and \GraphMLP, most methods improve the performance on the citation graphs, while for GAT the results are mixed. 
For the co-purchase dataset Computers and 622 split, all flat minima methods improve the results, while on the ra-pl split most of the methods decrease the performance.
We make a similar observation for the Photo dataset, but this time the ra-pl split is improved by the flat minima methods, except for four methods in combination GAT.
On PPI, the flat minima methods improve the results for GAT while the improvement is mixed for the other GNNs.

When comparing the flat minima methods overall, we notice that the methods extending SAM, \ie ASAM, PGN, and GSAM, do not consistently improve the results more than SAM. 
In most cases, one of the extensions works better than the original SAM, but this depends on the GNN and datasets.
For example, for GAT on the small datasets, SAM does on average not change the performance compared to the base model while ASAM reduces it by $0.22$ and PGN increases it by $0.25$ points.
Using ASAM instead of SAM for the adversarial calculation for PGN and GSAM works sometimes better, even though by itself SAM worked better than ASAM.
EWA works better than SWA with the highest improvement in the transductive setting of $2.09$ points when using EWA with a GCN on the CiteSeer and ra-pl split (see Table~\ref{tab:trans}).
This is likely because early stopping negatively impacts SWA.
SWA's begin and end epoch heavily depend on the model and dataset.
EWA nearly always begins in epoch $3$, ends one epoch after early stopping triggered, and in most cases uses a low $\alpha$ of $0.5$ to $0.9$ that favors more recent weights.

Anti-PGD works surprisingly well for a method that just adds noise to the model.
It is usually not the best method but, \eg on the small datasets with GAT it outperforms SAM, ASAM, SWA, EWA, and SAF.
It also reaches the overall highest accuracy on arXiv, which is achieved when applied to GCN.
While SAF is motivated by SAM, it impacts the models' performance differently.
For the small datasets with GCN it is worse than all the SAM-based methods, while with \GraphMLP it is better than all the SAM-based methods.
On arXiv, SAF is the only method that improves GAT's performance.
On PPI, SAF decrease the performance, while the SAM-based methods always improve it. 
The training of GAT with SAF on PPI was unstable and occasionally needed restarts.
SAF's recommended $\lambda = 0.3$ works for the citation datasets.
However, on PPI with GCN and GAT using $\lambda = 0.3$, early stopping is triggered at epoch $4\pm0$, \ie once SAF starts the performance only decreases.
Training is feasible with lower values, with $\lambda = 0.03$ GCN+SAF is the best model on PPI.

To the best of our knowledge, the only work who also applied flat minima methods to GNNs is the study by Kaddour et al.~\cite{Shchur18pitfalls}.
As said in the introduction, their work is limited to two flat minimal methods SAM and SWA and they also consider only one fixed train/test split.
We argue that it is crucial to consider randomized splits for a fair evaluation of flat minima methods on GNNs.
Considering the findings of \cite{kaddour2022questions}, they found that SAM works better than SWA on GNNs and that the results are influenced by the dataset and GNN architecture.
In contrast, we found that SAM works better than SWA.
This may be due to the additional randomized splits or the use of early stopping which affects SWA more than other methods.
Another reason may be that \cite{kaddour2022questions} used the original hyperparameter search space of SAM~\cite{Foret21sharpness}, while we additionally consider lower and higher values of $\rho$. 

For GCN one should use GASAM, for GAT one should use PGN, and for \GraphMLP one should use SAF.
In any case, one should always run one of the weight averaging methods as they do not need additional gradient computations.
In addition, one always obtains the original model without the modifications from the flat minima methods as well.
Finally, in our experiments we use early stopping.
Thus, it is important to decide on the hyperparameter values when to begin and stop averaging in SWA.
This choice of the hyperparameter is easier for EWA, since we start at a fixed epoch to average the weights.
Our hyperparameter search showed that EWA works well when one begins to average soon after the training starts and ending it when early stopping triggers, while using a strong decay value of $0.5$.
For \GraphMLP, SWA is the preferred flat minima method.
The reasons is that \GraphMLP trains for more epochs and SWA can better adapt the parameter weights.

\subsection{Combining up to Three Flat Minima Methods}
Above we mostly study existing methods, consider with EWA a variant of SWA, and with GASAM and PGNA combinations of two flat minima methods.
As proof of concept, we also further combine different flat minima methods without additional hyperparameter tuning.
As basis we use GCN in the transductive setting.
Table~\ref{tab:comb} shows the results for GCN with combinations of up to three flat minima methods with standard deviations in Table~\ref{tab:combsd}.
This shows that combining methods can increase the performance even further.
For example, on the CiteSeer ra-pl split, combining EWA and GASAM, which is a combination of EWA+GSAM+ASAM, increases the performance by $2.89$ points.

\begin{table*}[!!ht]
    \centering
    \caption{Transductive GCN with combination of up to three flat minima methods. SDs in tables \ref{tab:transSD} and \ref{tab:combsd}.}
    \label{tab:comb}
    \scriptsize
        \resizebox{\textwidth}{!}{

    \begin{tabular}{l|r r r | r r r | r r r | r r | r r }\toprule
    Dataset   & \multicolumn{3}{c|}{Cora} & \multicolumn{3}{c|}{CiteSeer} & \multicolumn{3}{c|}{PubMed} & \multicolumn{2}{c|}{Computer} & \multicolumn{2}{c}{Photo} \\
    Split     & plan & ra-pl & 622 & plan & ra-pl & 622 & plan & ra-pl & 622 & ra-pl & 622 & ra-pl & 622 \\ \midrule

  GCN       & $82.02$  & $79.82$  & $88.44$  & $71.39$  & $67.41$  & $76.81$  & $79.34$  & $77.27$  & $89.46$ & $82.78$ & $91.88$ & $90.89$ & $94.55$ \\
 + PGNA     & \good{$+0.35$} & \good{$+0.70$} & \bad{$-0.01$} & \good{$+0.93$} & \good{$+1.78$} & \bad{$-0.12$} & \good{$\mathbf{+0.22}$} & \good{$+0.33$} & \bad{$-0.02$} & \good{$+0.11$} & \good{$+0.06$} & \good{$+0.28$} & \good{$\mathbf{+0.02}$}\\
 + GASAM    & \good{$+0.35$} & \good{$+0.84$} & \good{$+0.02$} & \good{$+1.16$} & \good{$+1.58$} & \bad{$-0.12$} & \good{$+0.13$} & \good{$\mathbf{+0.43}$} & \bad{$-0.01$} & \bad{$-0.35$} & \good{$+0.10$} & \good{$+0.38$} & \bad{$-0.02$} \\ \hline
 + Anti-PGD+SAM  & \good{$+0.16$} & \good{$+0.78$} & \good{$+0.02$} & \good{$\mathbf{+1.40}$} & \good{$+1.60$} & \bad{$-0.11$} & \bad{$-0.15$} & \bad{$-0.06$} & \good{$+0.14$} & \good{$+0.13$} & \good{$+0.24$} & \good{$+0.43$} & \good{$+0.00$} \\
 + Anti-PGD+GASAM & \good{$\mathbf{+0.41}$} & \good{$+0.72$} & \good{$+0.02$} & \good{$+1.03$} & \good{$+1.89$} & \bad{$-0.12$} & \good{$+0.12$} & \good{$+0.32$} & \good{$+0.24$} & \bad{$-0.54$} & \good{$+0.25$} & \good{$+0.33$} & \good{$+0.01$} \\
 + Anti-PGD+SAF  & \good{$+0.02$} & \good{$\mathbf{+0.96}$} & \good{$+0.03$} & \good{$+0.26$} & \good{$+1.13$} & \bad{$-0.08$} & \good{$+0.08$} & \good{$+0.03$} & \good{$+0.21$} & \good{$\mathbf{+1.04}$} & \good{$+0.14$} & \good{$\mathbf{+0.62}$} & \bad{$-0.02$} \\
 + EWA+Anti-PGD  & \bad{$-0.09$} & \good{$+0.56$} & \good{$+0.03$} & \good{$+0.24$} & \good{$+1.14$} & \good{$\mathbf{+0.26}$} & \good{$+0.04$} & \bad{$-0.27$} & \good{$\mathbf{+0.32}$} & \bad{$-0.33$} & \good{$+0.26$} & \good{$+0.10$} & \good{$\mathbf{+0.02}$} \\
 + EWA+SAM  & \bad{$-0.32$} & \good{$+0.54$} & \good{$\mathbf{+0.06}$} & \good{$+0.68$} & \good{$+1.88$} & \good{$+0.08$} & \bad{$-0.34$} & \good{$+0.12$} & \good{$+0.00$} & \bad{$-0.19$} & \good{$+0.34$} & \good{$+0.35$} & \good{$+0.00$} \\
 + EWA+GASAM & \bad{$-0.50$} & \good{$+0.64$} & \good{$+0.00$} & \good{$+0.41$} & \good{$\mathbf{+2.89}$} & \good{$+0.08$} & \good{$+0.01$} & \good{$+0.35$} & \good{$+0.21$} & \bad{$-0.56$} & \good{$+0.32$} & \good{$+0.37$} & \bad{$-0.01$}  \\
 + EWA+SAF  & \good{$+0.12$} & \good{$+0.76$} & \good{$+0.03$} & \good{$+0.29$} & \good{$+2.06$} & \good{$\mathbf{+0.26}$} & \good{$+0.01$} & \bad{$-0.14$} & \good{$+0.24$} & \good{$+0.72$} & \good{$\mathbf{+0.38}$} & \good{$+0.56$} & \bad{$-0.02$} \\

    \bottomrule
    \end{tabular}
    }
\end{table*}

\subsection{Influence of Dataset Splits}
Regarding the dataset splits, we can see that the random split ra-pl is more difficult than the often used Planetoid split.
The Planetoid split uses a fixed set of 20 vertices per class as training data.
This makes models more susceptible to overfit the hyperparameters to that specific split than for a randomized ra-pl split.
For the 622 splits, the much higher amount of training data explain the overall better result.
Especially for the PubMed dataset, the amount of training data is larger by a factor of $200$ in the 622 split compared to the Planetoid split. 
In both the transductive and inductive settings, the dataset split impacts the performance one can gain from the flat minima methods.
The increase is on average higher and more consistent on the hardest ra-pl split compared to the other splits.

Shchur et al.~\cite{Shchur18pitfalls} show that randomized splits need to be used for a fair evaluation of GNNs.
We follow their suggestion by using two variants of randomized data splits. 
We confirm their observations that the commonly used ``Planetoid'' split is biased and should not be used on its own.
We extend on this observation and conclude from our experiments that randomized splits are also important for a fair evaluation of the flat minima methods applied to GNNs.

\subsection{Transductive vs. Inductive Training}
The hyperparameters and basic model performance are similar between the transductive and inductive setting.
In most cases the basic performance is within $1$ point.
The ranking of the flat minima methods is similar as well, \ie in many cases the best transductive method also works well in the inductive setting.
The major exception to this is \GraphMLP, discussed below.

\subsection{Detailed Discussion of Graph-MLP}
Compared to the original \GraphMLP~\cite{graphMLPwoMP} our modifications improved it by over $1$ point on Cora and CiteSeer and over $2$ points on PubMed.
The reason is that the original hyperparameter optimization was suboptimal.
For example, we found a larger batch/sample size to be beneficial. 
On the arXiv dataset, we found $\tau$ to be a critical hyperparameter and setting it outside the recommended $[.5, \dots, 2]$ range to $15$ increased the performance by roughly $5$ points.
On PPI, \GraphMLP completely fails.
The main reason for this is PPI's inductive nature, which means that \GraphMLP never uses the validation and testing edges.
This also explains the performance drop of \GraphMLP when we compare the same dataset between the transductive and inductive setting.
The size of the performance drop probably corresponds to the importance of knowing the edges that include testing vertices.
These edges seem to be very important for PPI, quite important for Cora where the performance dropped by over $10$ points, and not very important for PubMed where the drop was smaller than $1$ point, which is similar to the other GNNs.
In the transductive case, information about the edges connected to test vertices is available in training through $\hat A^r$.

\subsection{Assumptions and Limitations}
We assume that our GNN models provide a fair foundation for evaluating the flat minima methods.
We optimized the hyperparameters and checked the performance of the GNN models with the original works.
GCN and \GraphMLP achieve a performance on all datasets better than the literature~\cite{Kipf17gcn,graphMLPwoMP}.
The performance of our GAT model is slightly lower on CiteSeer, within standard error on Cora and PubMed, and better on PPI, compared with \cite{Velickovic18gat}.
Depending on the hyperparameters, the training of some GAT models on PPI was unstable and required multiple restarts. 

Our study considers the task of vertex classification.
We cover most nuances like small to large datasets with fixed and random splits, training in transductive and inductive settings, and single- and multi-label classification.
There are larger datasets than arXiv, but it is computationally very expensive to tune hyperparameters on these dataset for the GNN models and many flat minima methods considered here.
Beyond vertex classification, future extensions could consider also other tasks such as edge prediction and graph classification.

\section{Conclusion}
\label{sec:conclusion}
Overall our results show that the choice of the best flat minima method depends on the GNN model used and dataset split.
For the realistic and challenging random split datasets (ra-pl, 622), the flat minima methods can improve the GNN model more than on a fixed dataset split.
Shchur et al.~\cite{Shchur18pitfalls} argue for the need of using such random splits to fairly evaluate GNN models.
We extend on this and argue that a realistic assessment of flat minima methods on graph models requires such an evaluation procedure as well.
We observe that combining up to three flat minima methods can even further improve the results.
We recommend to always use weight averaging as SWA and EWA do not need any additional gradient calculations while also producing the original, unmodified models.
When using early stopping, we especially recommend using EWA.

\bibliographystyle{splncs04}
\bibliography{mybibliography}
\clearpage

\section*{Appendix}
\appendix

\section{Hyperparameters}
\label{appx:hyper}
We present the searched and final hyperparameters in this section.
For all experiments, Adam optimizer with PyTorch's default values $\beta_1 = 0.9$, $\beta_2 = 0.999$, and $eps=1e-08$ is used.
Early stopping with a patience of $100$ epochs and $20,000$ max epochs is applied.
After pre-experiments, we fixed the learning rates to the respective values and adjacency (edge) dropout to $0$ as well as all parameters not noted in the grid search ranges below.
Unless otherwise noted (\GraphMLP with arXiv and PPI), we did a full grid search over all combinations of the listed hyperparameters.
All final values are reported in Tables~\ref{tab:gcn_params}, \ref{tab:gcn_params_ind} for GCN, in Tables~\ref{tab:gat_params}, \ref{tab:gat_params_ind} for GAT, and in Tables~\ref{tab:gmlp_params}, \ref{tab:gmlp_params_ind} for \GraphMLP.

\begin{table*}[!hb]
    \centering   
        \caption{Optimal hyperparameter values for GCN on transductive tasks.}
    \label{tab:gcn_params}
    \tiny
    \begin{tabular}{l|r r r | r r r | r r r | r r | r r| r }\toprule
    Dataset       & \multicolumn{3}{c|}{Cora} & \multicolumn{3}{c|}{CiteSeer} & \multicolumn{3}{c|}{PubMed} & \multicolumn{2}{c|}{Computer} & \multicolumn{2}{c|}{Photo} & arXiv \\
    Split         & pl & ra-pl & 622 & pl & ra-pl & 622 & pl & ra-pl & 622 & ra-pl & 622 & ra-pl & 622 &- \\ \midrule
    input dropout & $0.15$ & $0.2$ & $0.0$ &  & $0.05$ &  & $0.2$ & $0.2$ & $0.0$ & $0.15$ & $0.15$ & $0.1$ & $0$ & $0.2$  \\
    model dropout & $0.8$ & $0.7$ & $0.4$ & $0.4$ & $0.6$ & $0.8$ & $0.5$ & $0.6$ & $0.8$ & $0.6$ & $0.8$ & $0.8$ & $0.5$ & $0.6$  \\
    weight decay  & $0.1$ & $0.1$ & $0.001$ & $0.316$ & $0.316$ & $0.01$ & $0.1$ & $0.01$ & $0.1$ & $0.01$ & $0.00316$ & $0.0316$ & $0.001$ & $0$ \\
    norm          & & id & & & id & & id & id & ln & ln & ln & ln & ln & ln \\
    residual con  & & no & & & no & & & no & & no & no & no & no & yes  \\
    num layers    & & $2$ & & & $2$ & & & $2$ & & $2$ & $2$ & $2$ & $2$ & $6$  \\
    hdim          & & $128$ & & & $256$ & & $128$ & $256$ & $128$ & $128$ & $256$ & $128$ & $256$ & $768$ \\
    lr            &  & $0.01$ &  &  & $0.01$ &  &  & $0.01$ &  & $0.01$ & $0.01$ & $0.01$ & $0.01$ & $0.001$  \\ \midrule
    SAM $\rho$    & $1$ & $1$ & $0.05$ & $5$ & $1$ & $5$ & $0.1$ & $0.5$ & $0.2$ & $2$ & $0.2$ & $5$ & $0.0005$ & $0.005$  \\
    ASAM $\rho$   & $10$ & $10$ & $0.5$ & $10$ & $20$ & $20$ & $0.1$ & $10$ & $0.01$ & $10$ & $0.5$ & $5$ & $0.02$ & $0.002$  \\
    PGN $\alpha$  & $0.3$ & $0.1$ & $0.3$ &  $0.4$ & $0.7$ &  $0.2$ &  $0.1$ & $0.3$ & $0.1$ & $0.4$ & $0.9$ & $0.3$ & $0.6$ & $0.2$  \\
    PGNA $\alpha$ & $0.3$ & $0.9$ & $0.3$ &  $0.5$ & $0.8$ &  $0.6$ &  $0.7$ & $0.5$ & $0.9$ & $0.1$ & $0.3$ & $0.6$ & $0.2$ & $0.9$  \\
    GSAM $\alpha$ & $0.5$ & $0.5$ & $5$ &  $0.5$ & $0.5$ &  $0.01$ & $0.002$ & $0.05$ & $0.002$ & $0.5$ & $1$ & $0.005$ & $0.5$ & $0.01$ \\
    GASAM $\alpha$ & $0.5$ & $1$ & $0.01$ &  $0.5$ & $1$ &  $0.5$ & $2$ & $0.005$ &  $2$ & $0.2$ & $0.5$ & $2$ & $5$ & $0.01$  \\
    SWA begin     & $75$ & $3$ & $3$ & & $3$ & & $75$ & $3$ & $77$ & $3$ & $75$ & $3$ & $25$ & $75$ \\
    SWA end       & $100$ & $1$ & $10$ & $1$ & $10$ & $10$ & $100$ & $25$ & $100$ & $100$ & $100$ & $10$ & $50$ & $100$  \\
    EWA begin     & & $3$ & & & $3$ & & & $3$ & & $3$ & $3$ & $3$ & $3$ & $3$  \\
    EWA end       & & $1$ & & $1$ & $1$ & $50$ & $1$ & $1$ & $100$ & $1$ & $100$ & $1$ & $1$ & $100$  \\
    EWA $\alpha$  & $0.5$ & $0.5$ & $0.5$ & $0.99$ & $0.99$ & $0.98$ & $0.8$ & $0.5$ & $0.9$ & $0.5$ & $0.95$ & $0.5$ & $0.5$ & $0.95$  \\
    Anti-PGD $\sigma$  & $0.003$ & $0.3$ & $0.003$ & $0.0003$ & $0.3$ & $0.001$ & $0.03$ & $0.03$ & $0.1$ & $0.03$ & $0.1$ & $0.01$ & $0.001$ & $0.001$  \\
    Anti-PGD $E$  & & $50$ & & $200$ & $50$ & $50$ & $50$ & $200$ & $200$ & $200$ & $200$ & $200$ & $200$ & $200$  \\
    SAF $\lambda$ & $0.5$ & $3$ & $0.1$ &  $3$ & $0.1$ &  $0.2$ & & $0.1$ & & $15$ & $3$ & $5$ & $0.3$ & $0.2$  \\
    SAF $\tau$    & $5$ & $2$ & $10$ &  & $5$ &  &  $10$ & $10$ & $2$ & $10$ & $5$ & $5$ & $2$ & $5$  \\
    \bottomrule
    \end{tabular}

\end{table*}

\begin{table*}[!hb]
    \centering   
        \caption{Optimal hyperparameter values for GCN on inductive tasks.}
    \label{tab:gcn_params_ind}
    \tiny
    \begin{tabular}{l|r r | r r | r r | r | r | r }\toprule
    Dataset       & \multicolumn{2}{c|}{Cora} & \multicolumn{2}{c|}{CiteSeer} & \multicolumn{2}{c|}{PubMed} & Computers & Photo & PPI\\
    Split         & pl & 622 & pl & 622 & pl & 622 & 622 & 622 & -\\ \midrule
    input dropout & $0.0$ & $0.15$ & $0.0$ & $0.15$ & $0.2$ & $0.05$ & $0.2$ & $0.2$ & $0.2$  \\
    model dropout & $0.8$ & $0.5$  & $0.8$ & $0.7$ & $0.5$ & $0.8$ & $0.7$ & $0.7$ & $0.4$  \\
    weight decay  & $0.001$ & $0.01$ & $0.316$ & $0.0316$ & $0.1$ & $0.01$ & $0.01$ & $0.001$ & $0.0001$  \\
    norm          & id & id & id & id & id & ln & ln & ln & ln  \\
    residual con           & no & no & no & no & no & no & no & no & yes  \\
    num layers    & $2$ & $2$ & $2$ & $2$ & $2$ & $2$ & $2$ & $2$ & $7$  \\
    hdim          & $256$ & $256$ & $256$ & $128$ & $128$ & $256$ & $256$ & $256$ & $2048$  \\
    lr            & $0.01$ & $0.01$ & $0.01$ & $0.01$ & $0.01$ & $0.01$ & $0.01$ & $0.01$ & $0.003$ \\ \midrule
    SAM $\rho$       & $2$ & $1$ & $5$ & $1$ & $0.0001$ & $2$ & $0.2$ & $0.001$ & $0.002$  \\
    ASAM $\rho$      & $20$ & $10$ & $20$ & $5$ & $0.005$ & $0.1$ & $0.01$ & $0.1$ & $0.001$ \\
    PGN $\alpha$   & $0.9$ & $0.3$ & $0.2$ & $0.4$ & $0.8$ & $0.1$ & $0.1$ & $0.9$ & $0.4$  \\
    PGNA $\alpha$   & $0.9$ & $0.4$ & $0.5$ & $0.5$ & $0.5$ & $0.1$ & $0.6$ & $0.2$ & $0.6$ \\
    GSAM $\alpha$   & $2$ & $0.05$ & $0.005$ & $0.1$ & $0.1$ & $0.0005$ & $0.2$ & $5$ & $0.005$   \\
    GASAM $\alpha$  & $2$ & $0.02$ & $1$ & $0.5$ & $0.02$ & $5$ & $0.01$ & $2$ & $0.002$  \\
    SWA begin     & $3$ & $3$ & $3$ & $3$ & $75$ & $100$ & $75$ & $25$ & $75$ \\
    SWA end       & $1$ & $50$ & $1$ & $25$ & $75$ & $100$ & $100$ & $100$ & $100$ \\
    EWA begin     & $3$ & $3$ & $3$ & $3$ & $3$ & $3$ & $3$ & $3$ & $3$ \\
    EWA end       & $1$ & $1$ & $1$ & $1$ & $1$ & $100$ & $100$ & $1$ & $100$ \\
    EWA $\alpha$  & $0.5$ & $0.8$ & $0.9$ & $0.9$ & $0.99$ & $0.9$ & $0.9$ & $0.5$ & $0.8$ \\
    Anti-PGD $\sigma$ & $0.03$ & $0.3$ & $0.1$ & $1$ & $0.001$ & $0.1$ & $0.01$ & $0.3$ & $0.03$ \\
    Anti-PGD $E$ & $50$ & $200$ & $50$ & $50$ & $50$ & $200$ & $50$ & $200$ & $200$ \\
    SAF $\lambda$   & $0.01$ & $10$ & $0.3$ & $0.3$ & $0.03$ & $0.3$ & $0.5$ & $1.5$ & $0.005$ \\
    SAF $\tau$      & $10$ & $10$ & $5$ & $5$ & $10$ & $5$ & $10$ & $2$ & $2$\\
    \bottomrule
    \end{tabular}

\end{table*}

\begin{table*}[!hb]
        \centering   
    \caption{Optimal hyperparameter values for GAT on transductive tasks.}
    \label{tab:gat_params}
        \tiny 
    \begin{tabular}{l|r r r | r r r | r r r | r r | r r | r }\toprule
    Dataset       & & Cora & & & CiteSeer & & & PubMed & & \multicolumn{2}{c|}{Computer} & \multicolumn{2}{c|}{Photo} & arXiv \\
    Split         & pl & rand pl & 622 & pl & rand pl & 622 & pl & rand pl & 622 & rp & 622 & rp & 622 &- \\ \midrule
    input dropout & $0.2$ & $0.2$ & $0.05$ & $0.05$ & $0.15$ & $0.15$ & $0.1$ & $0.15$ & $0.0$ & $0.2$ & $0.2$ & $0.15$ & $0.15$ & $0.2$  \\
    model dropout & $0.8$ & $0.7$ & $0.7$ & $0.8$ & $0.7$ & $0.8$ & $0.8$ & $0.7$ & $0.7$ & $0.4$ & $0.5$ & $0.4$ & $0.4$ & $0.5$  \\
    weight decay  & $0.0316$ & $0.01$ & $0.00316$ & $0.0316$ & $0.1$ & $0.01$ & $0.1$ & $0.1$ & $0.001$ & $0.01$ & $0.001$ & $0.001$ & $0.001$ & $0.0001$ \\
    norm          & ln & id & id & ln & id & id & & ln & & ln & ln & id & ln & bn \\
    residual con           & & no & & & no & & & no & & no & no & no & no & yes  \\
    num layers    & & $2$ & & & $2$ & & & $2$ & & $2$ & $2$ & $2$ & $2$ & $6$ \\
    hdim          & $32$ & $16$ & $32$ & $16$ & $32$ & $32$ & $32$ & $32$ & $32$ & $16$ & $32$ & $16$ & $32$ & $120$  \\
    lr            &  & $0.01$ &  &  & $0.01$ &  &  & $0.01$ &  & $0.01$ & $0.01$ & $0.01$ & $0.01$ & $0.001$  \\
    attn dropout  & & $0.5$ & & $0.5$ & $0.3$ & $0.5$ & $0.4$ & $0.3$ & $0.4$ & $0.3$ & $0.4$ & $0.4$ & $0.5$ & $0$  \\
    num attn head & & $8$ & & & $8$ & & & $8$ & & $8$ & $8$ & $8$ & $8$ & $3$  \\ \midrule
    SAM $\rho$    & $1$ & $0.001$ & $2$ & $2$ & $5$ & $2$ & $0.5$ & $0.5$ & $0.2$ & $0.5$ & $0.5$ &$0.001$ & $1$ & $0.05$  \\
    ASAM $\rho$   & $5$ & $20$ & $20$ &  & $10$ &  & $2$ & $5$ & $0.001$ & $2$ & $2$ & $0.1$ & $2$ & $0.1$  \\
    PGN $\alpha$  & $0.7$ & $0.5$ & $0.4$ &  $0.4$ & $0.1$ &  $0.1$ &  $0.8$ & $0.8$ & $0.9$ & $0.1$ & $0.5$ & $0.6$ & $0.3$ & $0.4$  \\
    PGNA $\alpha$ & $0.9$ & $0.5$ & $0.1$ &  $0.4$ & $0.4$ &  $0.2$ &  $0.4$ & $0.9$ & $0.9$ & $0.8$ & $0.5$ & $0.4$ & $0.9$ & $0.6$ \\
    GSAM $\alpha$ & $1$ & $0.2$ & $0.2$ &  $1$ & $0.01$ &  $0.1$ &  $2$ & $1$ &  $2$ & $1$ & $0.01$ & $1$ & $0.5$ & $0.002$  \\
    GASAM $\alpha$& $2$ & $0.5$ & $0.1$ &  $0.01$ & $2$ &  $1$ &  $5$ & $2$ &  $5$ & $0.1$ & $0.1$ & $0.002$ & $1$ & $0.02$  \\
    SWA begin     & $75$ & $3$ & $3$ & $75$ & $25$ & $25$ & $75$ & $3$ & $75$ & $75$ & $75$ & $3$ & $75$ & $3$ \\
    SWA end       & $100$ & $50$ & $10$ & $100$ & $50$ & $25$ & $100$ & $1$ & $100$ & $100$ & $100$ & $50$ & $100$ & $1$ \\
    EWA begin     & & $3$ & & & $3$ & & & $3$ & & $3$ & $3$ & $3$ & $3$ & $3$  \\
    EWA end       & & $1$ & & & $1$ & & $1$ & $1$ & $10$ & $1$ & $100$ & $1$ & $1$ & $1$ \\
    EWA $\alpha$  & & $0.5$ & & $0.5$ & $0.5$ & $0.9$ & $0.5$ & $0.5$ & $0.9$ & $0.5$ & $0.8$ & $0.5$ & $0.8$ & $0.99$  \\
    Anti-PGD $\sigma$  & $0.01$ & $0.003$ & $0.01$ & $0.0003$ & $0.1$ & $0.1$ & $0.001$ & $0.01$ & $0.1$ & $0.0003$ & $0.003$ & $0.003$ & $0.001$ & $0.01$  \\
    Anti-PGD $E$    & $200$ & $200$ & $50$ & $200$ & $50$ & $50$ & $50$ & $200$ & $200$ & $200$ & $50$ & $50$ & $200$ & $50$  \\
    SAF $\lambda$   & $0.1$ & $0.1$ & $0.2$ &  $0.3$ & $0.2$ &  $0.2$ & & $0.1$ & & $0.2$ & $0.1$ & $0.3$ & $0.01$ & $0.3$ \\
    SAF $\tau$      & & $10$ &  &  & $10$ &  &  $10$ & $2$ & $10$ & $10$ & $2$ & $10$ & $5$ & $2$  \\
    \bottomrule
    \end{tabular}
\end{table*}

\begin{table*}[!ht]
    \centering  
        \caption{Optimal hyperparameter values for GAT on inductive tasks.}
    \label{tab:gat_params_ind}
    \tiny
    \begin{tabular}{l|r r | r r | r r | r | r | r }\toprule
    Dataset       & \multicolumn{2}{c|}{Cora} & \multicolumn{2}{c|}{CiteSeer} & \multicolumn{2}{c|}{PubMed} & Computers & Photo & PPI\\
    Split         & pl & 622 & pl & 622 & pl & 622 & 622 & 622 & -\\ \midrule
    input dropout & $0.15$ & $0.05$ & $0.0$ & $0.1$ & $0.1$ & $0.1$ & $0.2$ & $0.15$ & $0$ \\
    model dropout & $0.8$ & $0.4$ & $0.6$ & $0.8$ & $0.8$ & $0.8$ & $0.5$ & $0.4$ & $0.1$ \\
    weight decay  & $0.0316$ & $0.01$ & $0.1$ & $0.01$ & $0.316$ & $0.001$ & $0.001$ & $0.001$ & $1E-6$ \\
    norm          & ln & ln & id & ln & ln & ln & ln & ln & id \\
    residual con           & no & no & no & no & no & no & no & no & yes \\
    num layers    & $2$  & $2$  & $2$  & $2$  & $2$  & $2$ & $2$ & $2$ & $7$  \\
    hdim          & $32$ & $32$ & $32$ & $32$ & $32$ & $32$ & $32$ & $32$ & $256$ \\
    lr            &$0.01$ &$0.01$ &$0.01$ &$0.01$ &$0.01$ &$0.01$ & $0.01$ & $0.01$ & $0.003$\\
    attn dropout  & $0.5$ & $0.5$ & $0.1$ & $0.4$ & $0.2$ & $0.1$ & $0.2$ & $0.2$ & $0.2$  \\
    num attn head & $8$   & $8$ & $8$ & $8$ & $8$ & $8$ & $8$ & $8$ & $8$ \\ \midrule
    SAM $\rho$    & $1$   & $0.5$ & $0.5$ & $1$ & $0.2$ & $1$ & $2$ & $0.5$ & $0.02$ \\
    ASAM $\rho$   &$0.002$& $1$ & $5$ & $10$ & $1$ & $0.5$ & $5$ & $5$ & $0.0005$ \\
    PGN $\alpha$  & $0.3$ & $0.9$ & $0.2$ & $0.2$ & $0.9$ & $0.9$ & $0.1$ & $0.5$ & $0.5$ \\
    PGNA $\alpha$ & $0.4$ & $0.1$ & $0.6$ & $0.4$ & $0.9$ & $0.1$ & $0.3$ & $0.2$ & $0.6$ \\
    GSAM $\alpha$ & $0.5$ &$0.005$& $0.05$ & $0.1$ & $0.02$ & $2$ & $0.05$ & $1$ & $0.002$ \\
    GASAM $\alpha$&$0.005$& $0.01$ & $0.01$ & $0.002$ & $2$ & $5$ & $0.02$ & $0.01$ & $0.002$ \\
    SWA begin     & $75$  & $75$ & $75$ & $3$ & $3$ & $75$ & $75$ & $75$ & $25$ \\
    SWA end       & $100$ & $100$ & $100$ & $50$ & $1$ & $100$ & $100$ & $100$ & $25$ \\
    EWA begin     & $3$   & $3$ & $75$ & $3$ & $3$ & $3$ & $75$ & $3$ & $3$  \\
    EWA end       & $1$   & $1$ & $100$ & $10$ & $1$ & $100$ & $100$ & $1$ & $1$ \\
    EWA $\alpha$  & $0.5$ & $0.5$ & $0.98$ & $0.95$ & $0.5$ & $0.95$ & $0.95$ & $0.5$ & $0.9$ \\
    Anti-PGD $\sigma$&$0.03$&$0.0003$& $0.1$ & $0.03$ & $0.01$ & $0.0003$ & $0.1$ & $0.001$ & $0.0003$ \\
    Anti-PGD $E$  & $50$ & $50$ & $50$ & $50$ & $200$ & $200$ & $50$ & $200$ & $50$ \\
    SAF $\lambda$ & $0.01$ & $0.07$ & $0.07$ & $0.2$ & $0.01$ & $0.02$ & $1$ & $0.1$ & $0.03$ \\
    SAF $\tau$    & $10$ & $2$ & $10$ & $2$ & $5$ & $10$ & $5$ & $10$ & $5$\\
    \bottomrule
    \end{tabular}

\end{table*}

\begin{table*}[!ht]
    \centering  
    \caption{Optimal hyperparameter values for \GraphMLP on transductive tasks.}
    \label{tab:gmlp_params}
    \tiny
    \begin{tabular}{l|r r r | r r r | r r r | r r | r r | r }\toprule
    Dataset       & & Cora & & & CiteSeer & & & PubMed & & \multicolumn{2}{c|}{Computer} & \multicolumn{2}{c|}{Photo} & arXiv \\
    Split         & pl & rand pl & 622 & pl & rand pl & 622 & pl & rand pl & 622 & rp & 622 & rp & 622 &- \\ \midrule
    input dropout & & $0$ & & & $0$ & & & $0$ & & $0$ & $0$ & $0$ & $0$ & $0$  \\
    model dropout & $0.4$ & $0.4$ & $0.6$ & $0.7$ & $0.8$ & $0.7$ & $0.5$ & $0.4$ & $0.2$ & $0.6$ & $0.3$ & $0.4$ & $0.5$ & $0.15$  \\
    weight decay  & $0.0001$ & $0.01$ & $0.001$ & $0.0001$ & $0.01$ & $0.0001$ & $0.001$ & $0.01$ & $0.001$ & $0.01$ & $0.001$ & $0.01$ & $0.001$ & $0.0$  \\
    norm          & & ln & & & ln & & & ln & & ln & ln & ln & ln & ln  \\
    residual con           & & no & & & no &  & & no & & no & no & no & no & yes \\
    num layers    & & $3$ & & & $3$ & & & $3$ & & $3$ & $3$ & $3$ & $3$ & $8$ \\
    hdim          & & $256$ & & & $256$ & & & $256$ & & $256$ & $256$ & $256$ & $256$ & $2048$ \\
    lr            &  & $0.01$ &  &  & $0.01$ &  &  & $0.01$ &  & $0.01$ & $0.01$ & $0.01$ & $0.01$ & $0.001$  \\
    NC @          & $-3$ & $-2$ & $-2$ & & $-2$ & & & $-2$ & & $-2$ & $-2$ & $-2$ & $-2$ & $-4$ \\
    NC weight     & $30$ & $1$ & $1$ & & $1$ & & & $1$ & & $10$ & $1$ & $3$ & $1$ & $30$ \\
    tau           & $0.5$ & $2$ & $2$ & $0.5$ & $2$ & $0.5$ & $2$ & $2$ & $1$ & $3$ & $10$ & $2$ & $10$ & $15$ \\
    r             & & $3$ & & & $3$ & & & $3$ & & $2$ & $2$ & $2$ & $2$ & $3$  \\
    b ($\%$ of data)& & $100$ & & & $100$ & & & $100$ & & $100$ & $100$ & $100$ & $100$ & $4$ \\ \midrule
    SAM $\rho$       & $0.5$ & $0.5$ & $0.005$ & $2$    & $1$ & $0.2$ & $0.05$ & $0.02$ & $0.05$ & $0.1$ & $0.01$ & $1$ & $5$ & $0.1$  \\
    ASAM $\rho$      & $2$   & $2$   & $0.5$   & $0.05$ & $5$ & $1$   & $0.2$ & $0.2$   & $0.01$ & $2$ & $0.0005$ & $5$ & $10$ & $0.05$  \\
    PGN $\alpha$   & $0.3$ & $0.2$ & $0.1$ &  $0.3$ & $0.3$ &  $0.5$ &  $0.1$ & $0.2$ & $0.5$ & $0.3$ & $0.9$ & $0.2$ & $0.1$ & $0.2$  \\
    PGNA $\alpha$   & $0.4$ & $0.7$ & $0.1$ & & $0.1$ & &  $0.5$ & $0.4$ &  $0.7$ & $0.1$ & $0.6$ & $0.2$ & $0.3$ & $0.4$  \\
    GSAM $\alpha$   & $0.01$ & $0.002$ & $1$ & $0.5$ & $0.1$ & $5.$ &  $0.005$ & $0.01$ &  $1$ & $2$ & $2$ & $0.002$ & $0.002$ & $0.01$  \\
    GASAM $\alpha$  & $0.1$ & $0.005$ & $2$ & $2$ & $0.01$ &  $5.$ &  $0.01$ & $0.2$ &  $0.01$ & $1$ & $0.5$ & $0.01$ & $0.005$ & $0.1$  \\
    SWA begin     & $25$ & $25$ & $75$ & $75$ & $75$ & $25$ & $75$ & $25$ & $75$ & $75$ & $75$ & $75$ & $75$ & $75$ \\
    SWA end       & $100$ & $50$ & $100$ & $100$ & $100$ & $50$ & $100$ & $25$ & $100$ & $100$ & $100$ & $100$ & $100$ & $100$ \\
    EWA begin     & & $3$ & & & $3$ & & & $3$ & & $3$ & $3$ & $3$ & $3$ & $3$  \\
    EWA end       & & $1$ & & & $1$ & & & $1$ & & $1$ & $1$ & $1$ & $1$ & $100$ \\
    EWA $\alpha$  & & $0.5$ & & & $0.5$ & & $0.5$ & $0.5$ & $0.8$ & $0.5$ & $0.8$ & $0.5$ & $0.5$ & $0.99$ \\
    Anti-PGD $\sigma$  & $0.01$ & $0.0003$ & $0.001$ & $0.003$ & $0.0003$ & $0.003$ & $0.01$ & $0.03$ & $0.1$ & $0.01$ & $0.03$ & $0.01$ & $0.1$ & $0.001$  \\
    Anti-PGD $E$    & & $200$ & & $200$ & $50$ & $50$ & $200$ & $200$ & $50$ & $50$ & $200$ & $50$ & $50$ & $200$  \\
    SAF $\lambda$   & $0.1$ & $2$ & $0.1$ & $0.1$ & $10$ & $0.1$ & $0.1$ & $0.1$ & $10$ & $0.5$ & $4$ & $0.07$ & $1$ & $0.3$  \\
    SAF $\tau$      & $10$ & $5$ & $5$ & $10$ & $10$ & $5$ & $10$ & $5$ & $5$ & $5$ & $5$ & $10$ & $2$ & $10$ \\
    \bottomrule
    \end{tabular}
\end{table*}

\begin{table*}[!ht]
    \centering   
    \caption{Optimal hyperparameter values for GMLP on inductive tasks.}
    \label{tab:gmlp_params_ind}
    \tiny
    \begin{tabular}{l|r r | r r | r r | r | r | r }\toprule
    Dataset       & \multicolumn{2}{c|}{Cora} & \multicolumn{2}{c|}{CiteSeer} & \multicolumn{2}{c|}{PubMed} & Computers & Photo & PPI\\
    Split         & pl & 622 & pl & 622 & pl & 622 & 622 & 622 & -\\ \midrule
    input dropout & $0$ & $0$ & $0$ & $0$ & $0$ & $0$ & $0$ & $0$ & $0$ \\
    model dropout & $0.8$ & $0.8$ & $0.8$ & $0.8$ & $0.6$ & $0.3$ & $0.6$ & $0.7$ & $0.1$ \\
    weight decay  & $0.1$ & $0.0001$ & $0.0001$ & $1E-5$ & $0.001$ & $0.01$ & $0.01$ & $0.01$ & $0.0001$ \\
    norm          & ln & ln & ln & ln & ln & ln & ln & ln & ln\\
    residual con           & no & no & no & no & no & no & no & no  & yes \\
    num layers    & $3$ & $3$ & $3$ & $3$ & $3$ & $3$ & $3$ & $3$ & $10$  \\
    hdim          & $256$ & $256$ & $256$ & $256$ & $256$ & $256$ & $256$ & $256$ & $2048$ \\
    lr            & $0.01$ & $0.01$ & $0.01$ & $0..01$ & $0.01$ & $0.01$ & $0.01$ & $0.01$ & $0.001$ \\ 
    NC@           & $-3$ & $-2$ & $-2$ & $-2$ & $-2$ & $-2$ & $-2$ & $-2$ & $-4$ \\
    NC weight     & $30$ & $1$ & $1$ & $1$ & $1$ & $1$ & $3$ & $1$ & $1$ \\
    tau           & $0.5$ & $1$ & $0.5$ & $0.5$ & $2$ & $1$ & $10$ & $5$ & $4$ \\
    r             & $3$ & $3$ & $3$ & $3$ & $3$ & $3$ & $2$ & $2$ & $3$ \\
     b ($\%$ of train)& $100$ & $100$ & $100$ & $100$ & $100$ & $100$ & $100$ & $100$ & $80$ \\\midrule
    SAM $\rho$     & $1$ & $5$ & $1$ & $5$ & $0.1$ & $0.5$ & $2$ & $1$ & $0.1$ \\
    ASAM $\rho$    & $10$ & $0.002$ & $0.02$ & $0.1$ & $0.5$ & $0.001$ & $10$ & $5$ & $0.5$ \\
    PGN $\alpha$   & $0.5$ & $0.4$ & $0.2$ & $0.2$ & $0.3$ & $0.5$ & $0.4$ & $0.2$ & $0.8$ \\
    PGNA $\alpha$  & $0.4$ & $0.1$ & $0.5$ & $0.1$ & $0.8$ & $0.7$ & $0.1$ & $0.2$ & $0.9$ \\
    GSAM $\alpha$  & $0.002$ & $0.02$ & $0.5$ & $0.002$ & $0.2$ & $2$ & $0.1$ & $0.01$ & $0.01$ \\
    GASAM $\alpha$ & $1$ & $5$ & $1$ & $5$ & $0.5$ & $0.05$ & $0.02$ & $0.05$ & $0.02$ \\
    SWA begin     & $75$ & $3$ & $75$ & $3$ & $75$ & $75$ & $75$ & $75$ & $75$ \\
    SWA end       & $100$ & $25$ & $100$ & $25$ & $100$ & $100$ & $100$ & $100$ & $100$ \\
    EWA begin     & $3$ & $3$ & $3$ & $3$ & $3$ & $3$ & $3$ & $3$ & $3$ \\
    EWA end       & $1$ & $1$ & $1$ & $1$ & $1$ & $1$ & $1$ & $1$ & $1$ \\
    EWA $\alpha$  & $0.8$ & $0.5$ & $0.5$ & $0.5$ & $0.5$ & $0.8$ & $0.5$ & $0.8$ & $0.5$ \\
    Anti-PGD $\sigma$ & $0.003$ & $0.01$ & $0.001$ & $0.03$ & $0.01$ & $0.3$ & $0.03$ & $0.1$ & $0.001$ \\
    Anti-PGD $E$ & $200$ & $50$ & $50$ & $200$ & $200$ & $50$ & $50$ & $200$ & $200$ \\
    SAF $\lambda$   & $7$ & $15$ & $1.5$ & $15$ & $0.5$ & $15$ & $0.4$ & $0.7$ & $0.07$ \\
    SAF $\tau$      & $2$ & $5$ & $10$ & $5$ & $10$ & $10$ & $5$ & $2$ & $10$  \\
    \bottomrule
    \end{tabular}
\end{table*}

\subsection{Base Models}
\paragraph{Small Datasets}
We use $20$ repeats/seeds for the pre-experiments and parameter search on the small datasets.
We ran pre-experiments to fix some parameters, and then ran a full grid search over the remaining parameter ranges.
For GCN, the searched space is input dropout in $\{.0, .05, .1, .15, .2\}$, model dropout in $\{.4, .5, .6, .7, .8\}$, weight decay in $\{.001, .00316, .01, .0316, .1, .316\}$, normalization in $\{id, ln\}$ ($id$ means no normalization, $ln$ layer norm, and $bn$ batch norm), and hidden dimension in $\{128, 256\}$. 
For Computer and Photo, model dropout was extended by  $\{.2, .3\}$.
For GAT, the number of attention heads is 8 in the first and 1 in the last layer.
The other parameters are searched with input dropout in $\{.0, .05, .1, .15, .2\}$, model dropout in $\{.4, .5, .6, .7, .8\}$, attention dropout in $\{.1, .2, .3, .4, .5\}$, weight decay in $\{.001, .00316, .01, .0316, .1, .316\}$, norm in $\{id, ln\}$, and hidden dimension in $\{16, 32\}$ (times 8 attention heads).
For Computer and Photo, weight decay was instead searched in $\{.0001, .000316, .001,\\ .00316, .01, .0316\}$.
Different to the original \GraphMLP we use dropout, layer norm, and activations between all layers.
The batch size $b$ to $100\%$ of each dataset.
The other parameters searched are model dropout in $\{.2, .3, .4, .5, .6, .7, .8\}$, weight decay in $\{ .1, .01, .001, .0001, 1E-5\}$, and $\tau$ in $\{.5, 1, 2\}$.
Loss weight and the layer after which the loss is calculated in $\{1@-2, 30@-3\}$, where layer $-1$ means after the last, $-2$ after the penultimate layer and so on.
For Photo and Computer model dropout was extended by $\{0, .1\}$, $\tau$ by $\{3, 5\}$, and the loss was instead searched in $\{0.3@-2, 1@-2, 3@-2, 10@-2\}$.

\paragraph{OGB arXiv}
We added reverse and self edges to the arXiv dataset which increased the accuracy by over $5$ points for most configurations.
We also used deeper models and added residual connections. 
We used $3$ repeats for the arXiv and PPI pre-experiments and parameter selection experiments.
For GCN, we searched input dropout in $\{.0, .1, .2, .3, .4\}$, model dropout in $\{.4, .5, .6, .7, .8\}$, and weight decay in $\{0, 1E-5, .0001\}$.
For GAT, we searched attention dropout in $\{.0, .1, .2, .3\}$, input dropout in $\{.0, .1, .2\}$, model drop-out in $\{.4, .5, .6, .7\}$, and weight decay in $\{0, .0001\}$.
For \GraphMLP, we did not perform a full grid search over the hyperparameters due to their larger number and \GraphMLP's lower training speed.
After fixing the other hyperparameters, we searched over many combinations of $b$ in $\{0.02, 0.04, 0.06, 0.08, 0.1, 1.2\}$, $NC@$ in $\{-2, -4, -6\}$, loss weight in $\{10, 30, 100\}$, $\tau$ in $\{0.5, 1, 1.5, 2, 2.5, 3, 5, 10, 15, 20, 25, 50, 100\}$ input dropout in $\{.0, .05\}$, and model dropout in $\{0.1, 0.15, 0.2, 0.25\}$.

\paragraph{PPI}
We used deeper models with residual connections for PPI as well. 
For PPI, the threshold to assign a label was chosen as $0.5$.
For GCN, we searched input dropout in $\{.0, .1, .2\}$, model dropout in $\{.2, .3, .4, .5, .6,\}$, and weight decay in $\{0, 1E-5, .0001\}$.
For GAT, we searched attention dropout in $\{.0, .1, .2\}$, input dropout in $\{.0, .1, .2\}$, model dropout in $\{.0, .1, .2, .3, .4\}$, and weight decay in $\{0, 1E-5, 1E-6\}$.
For \GraphMLP, we searched with drop input in $\{0, .1\}$, model dropout in $\{0, .1, .2, .3\}$, weight decay in $\{3E-5, .0001, 0.0003\}$, loss weight in $\{10,100\}$, and $NC@$ in $\{-4, -6, -8\}$, and $\tau$ in $\{3, 4, 5\}$.
Afterwards we searched for $b$ and found $0.8$ (\ie $80\%$ of each graph) to be the best value.

\subsection{Flat Minima Methods}

\paragraph{(A)SAM}
Both SAM and ASAM have the parameter $\rho$ which is usually set higher for ASAM than for SAM~\cite{Kwon21asam}, so we search over $\rho$ in $\{0.0001, 0.0002, 0.0005, 0.001,\\ 0.002, 0.005, 0.01, 0.02, 0.05, 0.1, 0.2, 0.5, 1, 2\}$ for SAM and $\rho$ in $\{0.0005, 0.001, \\ 0.002, 0.005, 0.01, 0.02, 0.05, 0.1, 0.2, 0.5, 1, 2, 5, 10\}$ for ASAM.
On the small data-sets, we saw potential for improvement with higher $\rho$ and thus additionally searched over $\{5, 10\}$ for SAM and $\{20, 50\}$ for ASAM.

\paragraph{PGN}
For PGN, we searched $\alpha$ in $\{0.1, 0.2, ..., 0.8, 0.9 \}$ in all cases.

\paragraph{GSAM}
For GSAM, we searched $\alpha$ in $\{0.002, 0.005, 0.01, 0.02, 0.05, 0.1, 0.2, 0.5, 1,\\ 2, 5\}$ for all models.

\paragraph{SWA and EWA}
For both SWA and EWA, we searched all combinations of $begin$ in $\{3, 25, 75\}$ and $end$ in $\{1, 10, 25, 50, 100\}$ where $end$ $>=$ $begin-3$.
This was done to prevent cases where no models are averaged at all as the lowest observed number of trained epochs from the first hyperparameter search is $5$.
For SWA, we averaged the model every epoch as we used fixed learning rates.
For EWA, we additionally tried the combinations above with $\alpha$ in $\{0.5, 0.8, 0.9, 0.95, 0.98, 0.99\}$.

\paragraph{Anti-PGD}
For Anti-PGD, we tried stopping the noise after $\{50, 200\}$ epochs and $\sigma$ in $\{0.0003, 0.001, \\ 0.003, 0.01, 0.03, 0.1, 0.3\}$.
For the small datasets, we additionally used $\sigma$ in $\{1, 3\}$.

\paragraph{SAF}
We always started SAF at epoch $5$ with a epoch difference $E$ of $3$.
We tested all combinations of $\tau$ in $\{2,5,10\}$ and $\lambda$ in $\{0.1, 0.2, 0.3, 0.4, 0.5, 0.7, 1, 2, 3\}$.
We noticed that the optimal $\lambda$ value often was on the border of that range so, we extended it by $\{0.01, 0.02, 0.03, 0.04, 0.05, 0.07\}$ on all datasets except arXiv, additionally by $\{1.5, 4, 5, 7, 10, 15\}$ on the small datasets, and additionally by $\{0.001, 0.002, 0.003, 0.005\}$ on PPI.

\section{Standard Deviations of Results}
Here we present the standard deviations of the main result tables for completeness, as they did not fit into the main tables but we still want to present them for interested readers.
Table \ref{tab:transSD} shows the standard deviations for the transductive results, Table \ref{tab:indSD} for the inductive results, and Table \ref{tab:combsd} for the combination of more flat minima methods.

\begin{table*}[!!ht]
    \centering
    \caption{SDs of Table \ref{tab:comb}, combined methods on transductive GCN.}
    \label{tab:combsd}
    \tiny
    \begin{tabular}{l|r r r | r r r | r r r | r r | r r }\toprule
    Dataset   & \multicolumn{3}{c|}{Cora} & \multicolumn{3}{c|}{CiteSeer} & \multicolumn{3}{c|}{PubMed} & \multicolumn{2}{c|}{Computer} & \multicolumn{2}{c}{Photo} \\
    Split     & plan & ra-pl & 622 & plan & ra-pl & 622 & plan & ra-pl & 622 & ra-pl & 622 & ra-pl & 622 \\ \midrule
 + Anti-PGD+SAM & $0.55$ & $1.35$ & $1.31$ & $0.76$ & $1.26$ & $1.44$ & $0.54$ & $2.23$ & $0.44$ & $1.88$ & $0.48$ & $0.90$ & $0.58$ \\
 + Anti-PGD+ASAM+GSAM & $0.41$ & $1.49$ & $1.33$ & $0.87$ & $1.50$ & $1.54$ & $0.44$ & $2.36$ & $0.47$ & $1.85$ & $0.47$ & $1.00$ & $0.58$ \\
 + Anti-PGD+SAF & $0.61$ & $1.50$ & $1.41$ & $1.63$ & $1.33$ & $1.51$ & $0.53$ & $2.12$ & $0.46$ & $1.94$ & $0.48$ & $1.08$ & $0.58$ \\
 + EWA+Anti-PGD & $0.61$ & $1.41$ & $1.37$ & $0.64$ & $1.39$ & $1.50$ & $0.33$ & $2.24$ & $0.48$ & $1.82$ & $0.47$ & $1.08$ & $0.56$ \\
 + EWA+SAM & $0.62$ & $1.38$ & $1.33$ & $1.38$ & $1.42$ & $1.46$ & $0.36$ & $2.16$ & $0.46$ & $2.06$ & $0.45$ & $0.96$ & $0.59$\\
 + EWA+ASAM+GSAM & $0.79$ & $1.59$ & $1.33$ & $0.64$ & $1.27$ & $1.46$ & $0.40$ & $2.22$ & $0.44$ & $2.03$ & $0.45$ & $0.96$ & $0.57$ \\
 + EWA+SAF & $0.65$ & $1.66$ & $1.34$ & $1.19$ & $1.26$ & $1.52$ & $0.38$ & $2.37$ & $0.44$ & $2.06$ & $0.48$ & $1.05$ & $0.59$ \\
    \bottomrule
    \end{tabular}
\end{table*}

\begin{table*}[!ht]
    \centering    
    \caption{Standard deviations for Table \ref{tab:trans}; 10 repeats on arXiv and 100 all other datasets}
    \label{tab:transSD}
    \tiny 
    \begin{tabular}{l| rrr | rrr | rrr | rr | rr | r}\toprule
    Dataset   & \multicolumn{3}{c|}{Cora} & \multicolumn{3}{c|}{CiteSeer} & \multicolumn{3}{c|}{PubMed} & \multicolumn{2}{c|}{Computers} & \multicolumn{2}{c|}{Photo} & arXiv \\
    Split     & plan & ra-pl & 622 & plan & ra-pl & 622 & plan & ra-pl & 622 & ra-pl & 622 & ra-pl & 622 & -\\ \midrule
  GCN    & $0.57$ & $1.58$ & $1.31$ & $1.20$ & $1.80$ & $1.57$ & $0.50$ & $2.28$ & $0.49$ & $1.85$ & $0.50$ & $1.19$ & $0.59$ & $0.12$ \\ \hline
 + SAM   & $0.50$ & $1.50$ & $1.34$ & $0.85$ & $1.66$ & $1.42$ & $0.45$ & $2.12$ & $0.43$ & $1.95$ & $0.49$ & $0.93$ & $0.57$ & $0.12$\\
 + ASAM  & $0.47$ & $1.48$ & $1.32$ & $0.86$ & $1.58$ & $1.53$ & $0.40$ & $2.16$ & $0.49$ & $1.89$ & $0.49$ & $0.97$ & $0.59$ & $0.11$\\
 + PGN   & $0.61$ & $1.43$ & $1.41$ & $1.08$ & $1.36$ & $1.53$ & $0.43$ & $2.25$ & $0.44$ & $1.94$ & $0.45$ & $0.87$ & $0.56$ & $0.14$\\
 + PGNA  & $0.54$ & $1.66$ & $1.39$ & $0.84$ & $1.69$ & $1.61$ & $0.42$ & $2.31$ & $0.43$ & $1.92$ & $0.48$ & $0.91$ & $0.55$ & $0.16$\\
 + GSAM  & $0.55$ & $1.25$ & $1.32$ & $0.89$ & $1.63$ & $1.47$ & $0.46$ & $2.13$ & $0.46$ & $1.93$ & $0.42$ & $0.97$ & $0.58$ & $0.14$\\
 + GASAM & $0.48$ & $1.33$ & $1.33$ & $0.85$ & $1.61$ & $1.49$ & $0.53$ & $2.22$ & $0.51$ & $1.88$ & $0.51$ & $0.99$ & $0.55$ & $0.13$\\ \hline
 + SWA   & $0.26$ & $1.40$ & $1.30$ & $0.40$ & $1.31$ & $1.46$ & $0.33$ & $2.36$ & $0.47$ & $2.00$ & $0.47$ & $1.01$ & $0.56$ & $0.12$\\
 + EWA   & $0.71$ & $1.53$ & $1.32$ & $0.61$ & $1.25$ & $1.51$ & $0.37$ & $2.30$ & $0.46$ & $2.00$ & $0.48$ & $1.25$ & $0.57$ & $0.07$\\ \hline
 + Anti-PGD & $0.62$ & $1.40$ & $1.40$ & $1.06$ & $1.36$ & $1.57$ & $0.54$ & $2.18$ & $0.45$ & $1.86$ & $0.58$ & $1.14$ & $0.56$ & $0.12$\\
 + SAF   & $0.57$ & $1.71$ & $1.36$ & $0.79$ & $1.97$ & $1.60$ & $0.58$ & $2.28$ & $0.49$ & $1.91$ & $0.46$ & $1.07$ & $0.57$ & $0.15$\\ \hline
 \midrule
  GAT   & $0.84$ & $1.34$ & $1.36$ & $0.82$ & $1.11$ & $1.52$ & $0.83$ & $2.33$ & $0.50$ & $1.90$ & $0.47$ & $1.32$ & $0.56$ & $0.15$\\ \hline
 + SAM  & $0.95$ & $1.28$ & $1.36$ & $1.01$ & $1.29$ & $1.57$ & $1.37$ & $2.61$ & $0.49$ & $1.87$ & $0.43$ & $1.28$ & $0.64$ & $0.11$\\
 + ASAM & $0.89$ & $1.26$ & $1.42$ & $0.97$ & $1.24$ & $1.59$ & $1.17$ & $2.44$ & $0.49$ & $1.95$ & $0.46$ & $1.29$ & $0.60$ & $0.10$\\
 + PGN  & $0.69$ & $1.36$ & $1.46$ & $0.87$ & $1.22$ & $1.50$ & $0.99$ & $2.40$ & $0.48$ & $1.76$ & $0.50$ & $1.19$ & $0.58$ & $0.12$\\
 + PGNA & $0.67$ & $1.50$ & $1.37$ & $0.77$ & $1.29$ & $1.51$ & $0.80$ & $2.34$ & $0.49$ & $1.93$ & $0.44$ & $1.21$ & $0.57$ & $0.11$\\
 + GSAM & $0.72$ & $1.31$ & $1.45$ & $0.93$ & $1.25$ & $1.45$ & $1.11$ & $2.55$ & $0.49$ & $1.75$ & $0.46$ & $1.24$ & $0.60$ & $0.14$\\
 + GASAM& $0.67$ & $1.36$ & $1.45$ & $0.95$ & $1.27$ & $1.57$ & $0.99$ & $2.21$ & $0.48$ & $1.86$ & $0.47$ & $1.25$ & $0.57$ & $0.16$\\ \hline
 + SWA  & $0.59$ & $1.32$ & $1.38$ & $0.44$ & $1.08$ & $1.44$ & $0.64$ & $2.45$ & $0.52$ & $2.04$ & $0.44$ & $1.30$ & $0.55$ & $5.26$\\
 + EWA  & $0.74$ & $1.24$ & $1.45$ & $0.74$ & $1.08$ & $1.47$ & $0.87$ & $2.31$ & $0.52$ & $1.97$ & $0.44$ & $1.27$ & $0.58$ & $6.77$\\ \hline
 + Anti-PGD & $0.99$ & $1.23$ & $1.37$ & $0.74$ & $1.06$ & $1.53$ & $0.85$ & $2.17$ & $0.48$ & $1.82$ & $0.47$ & $1.25$ & $0.61$ & $0.16$\\
 + SAF  & $0.87$ & $1.38$ & $1.36$ & $0.75$ & $1.14$ & $1.50$ & $0.87$ & $2.32$ & $0.49$ & $1.81$ & $0.45$ & $1.21$ & $0.54$ & $0.11$\\ \hline
\midrule
  \GraphMLP & $0.68$ & $1.65$ & $1.25$ & $0.60$ & $1.26$ & $1.57$ & $0.86$ & $2.29$ & $0.42$ & $2.07$ & $0.45$ & $1.30$ & $0.51$ & $0.50$\\ \hline
 + SAM & $0.66$ & $1.75$ & $1.29$ & $0.77$ & $1.01$ & $1.47$ & $0.83$ & $2.31$ & $0.44$ & $2.05$ & $0.46$ & $1.11$ & $0.52$  & $0.46$\\
 + ASAM & $0.68$ & $1.68$ & $1.37$ & $0.64$ & $1.24$ & $1.61$ & $0.80$ & $2.34$ & $0.44$ & $1.91$ & $0.48$ & $1.01$ & $0.47$ & $0.61$\\
 + PGN & $0.64$ & $1.72$ & $1.41$ & $0.60$ & $1.01$ & $1.51$ & $0.77$ & $2.59$ & $0.47$ & $2.02$ & $0.45$ & $1.00$ & $0.49$ & $0.36$\\
 + PGNA & $0.80$ & $1.75$ & $1.29$ & $0.65$ & $1.16$ & $1.51$ & $0.78$ & $2.30$ & $0.50$ & $1.98$ & $0.46$ & $0.91$ & $0.48$ & $0.32$\\
 + GSAM & $0.74$ & $1.78$ & $1.26$ & $0.77$ & $1.41$ & $1.45$ & $0.75$ & $2.33$ & $0.44$ & $1.87$ & $0.46$ & $1.23$ & $0.44$ & $0.35$\\
 + GASAM & $0.67$ & $1.67$ & $1.33$ & $0.65$ & $1.11$ & $1.37$ & $0.75$ & $2.35$ & $0.44$ & $1.98$ & $0.47$ & $1.00$ & $0.47$ & $0.48$\\ \hline
 + SWA & $0.48$ & $1.23$ & $1.31$ & $0.55$ & $0.98$ & $1.57$ & $0.73$ & $2.50$ & $1.03$ & $1.86$ & $0.48$ & $1.08$ & $0.49$ & $0.24$\\
 + EWA & $0.65$ & $1.54$ & $1.24$ & $0.58$ & $1.22$ & $1.57$ & $0.96$ & $2.28$ & $0.41$ & $2.08$ & $0.46$ & $1.35$ & $0.49$ & $0.31$\\ \hline
 + Anti-PGD & $0.79$ & $1.76$ & $1.25$ & $0.65$ & $1.14$ & $1.48$ & $0.74$ & $2.30$ & $0.37$ & $1.98$ & $0.47$ & $1.37$ & $0.50$ & $0.42$\\
 + SAF & $0.67$ & $1.59$ & $1.41$ & $0.62$ & $1.28$ & $1.40$ & $0.72$ & $2.33$ & $0.46$ & $1.83$ & $0.50$ & $1.38$ & $0.45$ & $0.23$\\
    \bottomrule
    \end{tabular}
\end{table*}

\begin{table*}[!ht]
    \centering
        \caption{Standard deviations for Table \ref{tab:ind}; 10 repeats on PPI and 100 all other datasets}
    \label{tab:indSD}
   \tiny
    \begin{tabular}{l|r r | r r | r r | r | r | r }\toprule
    Dataset       & \multicolumn{2}{c|}{Cora} & \multicolumn{2}{c|}{CiteSeer} & \multicolumn{2}{c|}{PubMed} & Computers & Photo & PPI\\
    Split         & plan & 622 & plan & 622 & plan & 622 & 622 & 622 & -\\ \midrule

  GCN & $0.46$ & $1.38$ & $0.99$ & $1.52$ & $0.59$ & $0.46$ & $0.48$ & $0.59$ & $0.03$\\
 + SAM   & $0.56$ & $1.35$ & $1.01$ & $1.65$ & $0.49$ & $0.47$ & $0.47$ & $0.59$ & $0.02$\\
 + ASAM  & $0.60$ & $1.42$ & $0.80$ & $1.59$ & $0.51$ & $0.51$ & $0.49$ & $0.60$ & $0.03$\\
 + PGN   & $0.50$ & $1.51$ & $0.89$ & $1.59$ & $0.57$ & $0.48$ & $0.51$ & $0.59$ & $0.02$\\
 + PGNA  & $0.60$ & $1.41$ & $0.99$ & $1.59$ & $0.52$ & $0.43$ & $0.48$ & $0.59$ & $0.04$\\
 + GSAM  & $0.50$ & $1.48$ & $1.16$ & $1.59$ & $0.46$ & $0.46$ & $0.49$ & $0.60$ & $0.02$\\
 + GASAM & $0.53$ & $1.35$ & $0.88$ & $1.65$ & $0.46$ & $0.46$ & $0.49$ & $0.62$ & $0.03$\\ \hline
 + SWA   & $0.46$ & $1.42$ & $0.25$ & $1.63$ & $0.46$ & $0.44$ & $0.49$ & $0.58$ & $0.06$\\
 + EWA   & $0.46$ & $1.33$ & $0.37$ & $1.63$ & $0.31$ & $0.47$ & $0.48$ & $0.59$ & $0.02$\\ \hline
 + Anti-PGD   & $0.51$ & $1.35$ & $0.65$ & $1.57$ & $0.50$ & $0.48$ & $0.52$ & $0.58$ & $0.03$ \\
 + SAF   & $0.47$ & $1.55$ & $0.97$ & $1.54$ & $0.53$ & $0.47$ & $0.53$ & $0.61$ & $0.01$\\ 

\hline
\midrule
  GAT & $1.14$ & $1.39$ & $0.48$ & $1.56$ & $0.84$ & $0.47$ & $0.50$ & $0.65$ & $0.03$\\
 + SAM & $0.81$ & $1.46$ & $0.53$ & $1.51$ & $0.94$ & $0.52$ & $0.45$ & $0.65$ & $0.03$\\
 + ASAM & $0.70$ & $1.38$ & $0.52$ & $1.64$ & $1.01$ & $0.49$ & $0.47$ & $0.65$ & $0.03$\\
 + PGN & $0.90$ & $1.47$ & $0.54$ & $1.62$ & $0.92$ & $0.51$ & $0.49$ & $0.61$ & $0.02$\\
 + PGNA & $1.00$ & $1.36$ & $0.46$ & $1.63$ & $0.94$ & $0.49$ & $0.49$ & $0.62$ & $0.02$\\
 + GSAM & $0.89$ & $1.40$ & $0.50$ & $1.53$ & $0.87$ & $0.56$ & $0.50$ & $0.65$ & $0.04$\\
 + GASAM & $0.74$ & $1.43$ & $0.50$ & $1.62$ & $1.01$ & $0.50$ & $0.47$ & $0.62$ & $0.03$\\ \hline
 + SWA & $0.56$ & $1.29$ & $0.17$ & $1.53$ & $0.96$ & $0.49$ & $0.50$ & $0.64$ & $0.73$\\
 + EWA & $1.01$ & $1.27$ & $0.23$ & $1.59$ & $0.84$ & $0.47$ & $0.47$ & $0.68$ & $0.05$\\ \hline
 + Anti-PGD & $0.79$ & $1.36$ & $0.30$ & $1.59$ & $0.87$ & $0.51$ & $0.53$ & $0.66$ & $0.04$\\
 + SAF & $1.10$ & $1.43$ & $0.48$ & $1.53$ & $0.87$ & $0.47$ & $0.48$ & $0.66$ & $0.12$ \\

\hline
\midrule
  \GraphMLP & $0.97$ & $1.66$ & $0.85$ & $1.48$ & $0.92$ & $0.48$ & $0.62$ & $0.61$ & $1.12$\\ \hline
 + SAM & $1.12$ & $1.62$ & $0.72$ & $1.53$ & $0.83$ & $0.41$ & $0.60$ & $0.57$ & $1.60$\\
 + ASAM & $1.25$ & $1.66$ & $0.81$ & $1.40$ & $0.81$ & $0.49$ & $0.61$ & $0.58$ & $1.80$\\
 + PGN & $1.19$ & $1.63$ & $0.77$ & $1.48$ & $0.86$ & $0.46$ & $0.62$ & $0.55$ & $0.26$\\
 + PGNA & $1.28$ & $1.53$ & $0.70$ & $1.40$ & $0.76$ & $0.44$ & $0.57$ & $0.58$ & $0.25$\\
 + GSAM & $1.40$ & $1.66$ & $0.78$ & $1.54$ & $0.87$ & $0.36$ & $0.57$ & $0.64$ & $1.47$\\
 + GASAM & $1.20$ & $1.68$ & $0.69$ & $1.54$ & $0.70$ & $0.47$ & $0.56$ & $0.59$ & $0.97$\\ \hline
 + SWA & $0.55$ & $1.54$ & $0.65$ & $1.44$ & $0.85$ & $0.51$ & $0.56$ & $0.54$ & $0.93$\\
 + EWA & $0.84$ & $1.67$ & $0.83$ & $1.47$ & $0.90$ & $0.42$ & $0.58$ & $0.61$ & $1.06$\\ \hline
 + Anti-PGD & $0.90$ & $1.55$ & $0.73$ & $1.42$ & $0.77$ & $0.51$ & $0.57$ & $0.61$ & $0.42$\\
 + SAF & $0.80$ & $1.53$ & $0.87$ & $1.54$ & $0.57$ & $0.39$ & $0.59$ & $0.61$ & $0.23$\\

   \bottomrule
    \end{tabular}
\end{table*}


\end{document}